\documentclass[runningheads]{llncs}
\usepackage{graphicx}

\usepackage{graphicx}
\usepackage{amsmath}
\usepackage{amssymb}
\usepackage{booktabs}
\usepackage[pagebackref,breaklinks,colorlinks]{hyperref}
\usepackage{adjustbox}
\usepackage{soul}
\usepackage{enumitem}
\usepackage{float}
\usepackage{cite}
\usepackage{enumitem}
\usepackage{caption}
\usepackage[capitalize]{cleveref}
\crefname{section}{Sec.}{Secs.}
\Crefname{section}{Section}{Sections}
\Crefname{table}{Table}{Tables}
\crefname{table}{Tab.}{Tabs.}

\usepackage{color}
\definecolor{darkgreen}{rgb}{0,0.5,0}
\definecolor{purple}{rgb}{1,0,1}
\newcommand{\kibitz}[2]{\ifnum\Comments=1\textcolor{#1}{#2}\fi}

\newcommand{\etal}{\textit{et al.}}
\newcommand{\eg}{\textit{e.g.}}
\newcommand{\ie}{\textit{i.e.}}
\newcommand\tb[1]{\textbf{#1}}

\usepackage{tikz}
\usepackage{comment}
\usepackage{amsmath,amssymb} 
\usepackage{color}

\begin{document}
\pagestyle{headings}
\mainmatter
\def\ECCVSubNumber{5361}  

\title{All You Need is RAW: \\Defending Against Adversarial Attacks with Camera Image Pipelines }

\author{
 \hspace{-0mm}
Yuxuan Zhang
\qquad
Bo Dong
\qquad
Felix Heide  \vspace{2mm}
}

\institute{Princeton University}

\titlerunning{Defending Against Adversarial Attacks with Camera Image Pipelines }
%
%
\authorrunning{Y.Zhang,  B.Dong,  F.Heide}
%
\maketitle

\begin{abstract}
    Existing neural networks for computer vision tasks are vulnerable to adversarial attacks: adding imperceptible perturbations to the input images can fool these models to make a false prediction on an image that was correctly predicted without the perturbation. Various defense methods have proposed image-to-image mapping methods, either including these perturbations in the training process or removing them in a preprocessing step. In doing so, existing methods often ignore that the natural RGB images in today's datasets are not captured but, in fact, recovered from RAW color filter array captures that are subject to various degradations in the capture. In this work, we exploit this RAW data distribution as an empirical prior for adversarial defense. Specifically, we proposed a model-agnostic adversarial defensive method, which maps the input RGB images to Bayer RAW space and back to output RGB using a learned camera image signal processing (ISP) pipeline to eliminate potential adversarial patterns. The proposed method acts as an off-the-shelf preprocessing module and, unlike model-specific adversarial training methods, does not require adversarial images to train. As a result, the method generalizes to unseen tasks without additional retraining. Experiments on large-scale datasets (\eg, ImageNet, COCO) for different vision tasks (\eg, classification, semantic segmentation, object detection) validate that the method significantly outperforms existing methods across task domains.

\vspace{-3mm}
\end{abstract}

\section{Introduction}

\vspace{-2mm}
The most successful methods for a broad range of tasks in computer vision rely on deep neural networks \cite{chen2017deeplab,he2017mask,he2016deep,Isola_2017_CVPR,zhang2016colorful} (DNNs), including classification, detection, segmentation, scene understanding, scene reconstruction and generative tasks. Although we rely on the predictions of DNNs in safety-critical applications in robotics, self-driving vehicles, medical diagnostics, and video security, existing networks have been shown to be vulnerable to adversarial attacks\cite{szegedy2014intriguing}: small perturbations to images that are imperceptible to the human vision system to images can deceive DNNs to make incorrect predictions\cite{madry2017towards,poursaeed2018generative,szegedy2013intriguing,nakkiran2019adversarial,tsipras2019robustness}. As such, defending against adversarial attacks\cite{borkar2020defending,lu2017safetynet,madry2017towards,papernot2016distillation,xie2019feature} can help resolve failure cases in safety-critical applications and provide insights into the generalization capabilities of training procedures and network architectures.


\begin{figure}[t]
\vspace{-0mm}
\begin{center}
\begin{minipage}{0.58\linewidth}
 \begin{adjustbox}{width=1\linewidth}
\includegraphics[width=1\linewidth, trim=0 0 0 0,clip]{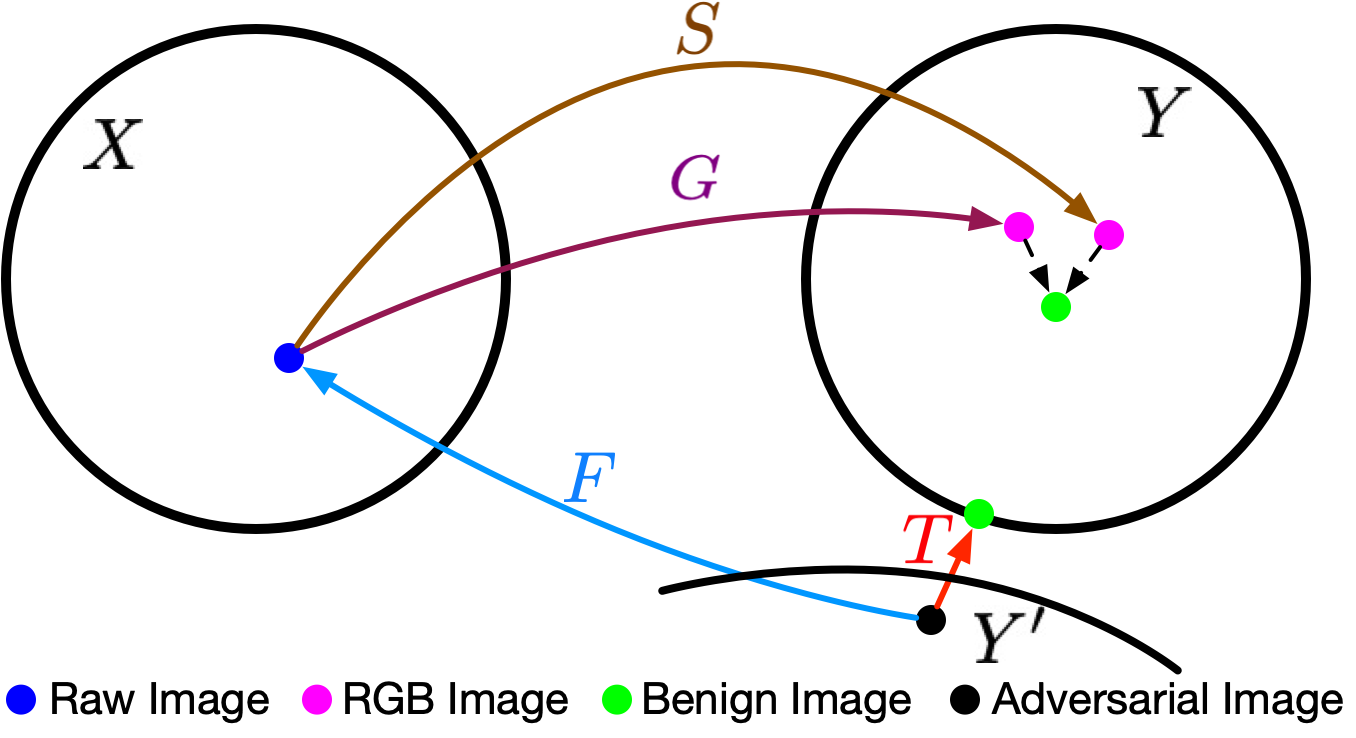}
\end{adjustbox}
\end{minipage}
\hfill
\begin{minipage}{0.4\linewidth}
\captionsetup{font=footnotesize,labelfont=footnotesize}
\caption{
{Existing defense approaches learn an RGB-to-RGB projection from an adversarial distribution ($Y'$) to its natural RGB distribution ($Y$): $T: Y' \rightarrow Y$. In contrast, our approach learns a mapping via the intermediate natural RAW distribution ($X$), which is achieved by utilizing three specially designed operators: $F: Y' \rightarrow X$, $G: X \rightarrow Y$, and $S: X \rightarrow Y$.
}
}
\label{fig:RawDistribution}
\end{minipage}
\end{center}
\vspace{-10mm}
\end{figure}

Existing defense methods fall into two approaches: They either introduce adversarial examples to the training dataset, resulting in new model weights, or they transform the inputs, aiming to remove the adversarial pattern, before feeding them into the unmodified target models. Specifically, the first line of defense methods generates adversarial examples by iteratively training a target model while finding and adding remaining adversarial images as training samples in each iteration\cite{yin2019gat, wong2020fast}  \cite{wong2020fast, wang2019bilateral, Gong_2021_CVPRMaxup}. Although the set of successful adversarial examples shrinks over time, iteratively generating them is extremely costly in training time, and different adversarial images must be included for defending against different attack algorithms. Moreover, the adversarial examples cannot be stored once in a training set as they are model-specific and domain-specific, meaning they must be re-generated when used for different models or on other domains. 

Defense methods that transform the input image aim to overcome the limitations of adversarial training approaches. Considering adversarial perturbations as noise, these methods ``denoise'' the inputs before feeding them into unmodified target models. The preprocessing module can either employ image-to-image models such as auto-encoders or generative adversarial methods\cite{jia2019comdefend, samangouei2018defense, liu2019feature},  or they rely on conventional image-processing operations\cite{guo2017countering, das2017keeping, jpeg2016, GuidedDenoiser}. Compared to adversarial training methods, these methods are model-agnostic and require no adversarial images for training. 

All methods in this approach have in \emph{common that they rely on RGB image data as input and output}. That is, they aim to recover the distribution of natural RGB images and project the adversarial image input to the closest match in this distribution, using a direct image-to-image mapping network. As such, existing methods often ignore the fact that images in natural image datasets are the result of several processing steps applied to the raw captured images. In particular, training image datasets are produced by interpolating sub-sampled, color filtered (\eg, using Bayer filter) raw data, followed by a rich low-level processing pipeline, including readout and photon noise denoising. As a result, the raw per-pixel photon counts are heavily subsampled, degraded and processed in an RGB image. We rely on the \emph{RAW data distribution, before becoming RGB images, as a prior in the proposed adversarial defense method}, which is empirically described in large datasets of RAW camera captures. Specifically, instead of directly learning a mapping between adversarially perturbed inputs RGBs and ``clean'' output RGBs, we learn a mapping via the intermediate RAW color filter array domain. In this mapping, we rely on learned ISP pipelines as low-level camera image processing blocks to map from RAW to RGB. The resulting method is entirely model-agnostic, requires no adversarial examples to train, and acts as an off-the-shelf preprocessing module that can be transferred to any task on any domain. We validate our method on large-scale datasets (ImageNet, COCO) for different vision tasks (classification, semantic segmentation, object detection), and also perform extensive ablation studies to assess the robustness of the proposed method to various attack methods, model architecture choices, and hyper-parameters choices.
\\

\noindent
Specifically, we make the following contributions:
\begin{itemize}
	\item[\textbullet] We propose, to the best of our knowledge, the first adversarial defense method that exploits the natural distribution of RAW domain images.
	\vspace{1mm}
	\item[\textbullet] The proposed method avoids the tedious generation of adversarial training images and can be used as an off-the-shelf preprocessing module for diverse tasks. 
	\vspace{1mm}
	\item[\textbullet] We provide a detailed analysis of how the natural RAW image distribution helps defend against adversarial attacks, and we validate that the method achieves \emph{state-of-the-art} defense accuracy for input transformation defenses, outperforming existing approaches.	
\end{itemize}
We will provide all code, models, and instructions needed to reproduce the results presented in this work.

\vspace{-5pt}
\section{Related Work}
\vspace{-2mm}
\subsection{Camera Image Signal Processing (ISP) Pipeline}
\vspace{-2mm}
A camera image signal processing (ISP) pipeline converts RAW measurements from a digital camera sensor to high-quality images suitable for human viewing or downstream analytic tasks. To this end, a typical ISP pipeline encompasses a sequence of modules~\cite{Karaimer2016ASP} each addressing a portion of this image reconstruction problem. In a hardware ISP, these modules are proprietary compute units, and their behavior is unknown to the user. More importantly, the modules are not differentiable~\cite{10.1145/3306346.3322996, isp_opt_cvpr20}. Two lines of works leveraged deep-learning-based approaches to cope with the significant drawback. 

One line of the works directly replaced the hardware ISP with a deep-learning-based model to target different application scenarios, such as low-light enhancement~\cite{DBLP:conf/cvpr/ChenCXK18, DBLP:conf/iccv/ChenCDK19}, super-resolution~\cite{8953691, xu2021exploiting, zhang2019zoom}, smartphone camera enhancement~\cite{10.1109/TIP.2018.2872858, Dai2020AWNetAW, Ignatov2020ReplacingMC}, and ISP replacement~\cite{9329084}. Nevertheless, the deep-learning-based models used by these works contain a massive number of parameters and are computationally expensive. Thus, their application is limited to off-line tasks. 
In contrast, another thread of works focused on searching for the best hardware ISP hyperparameters for different downstream tasks, by leveraging deep-learning-based approaches. 
Specifically, Tseng~\etal~\cite{10.1145/3306346.3322996} proposed differentiable proxy functions to model arbitrary ISP pipelines and leveraged them to find the best hardware ISP hyperparameters for different downstream tasks. Yu~\etal~\cite{Yu2021ReconfigISPRC} proposed ReconfigISP, which uses different proxy functions for each module of a hardware ISP instead of the whole ISP pipeline. Mosleh~\etal~\cite{isp_opt_cvpr20} proposed a hardware-in-the-loop method to optimize hyperparameters of a hardware ISP directly.

\vspace{-5mm}
\subsection{Adversarial Attack Methods}
\vspace{-2mm}
Adversarial attacks have drawn significant attention from the deep-learning community. Based on the access level to target networks, adversarial attacks can be broadly divided into white-box attacks and black-box attacks.

Among the white-box attack, one important stream is gradient-based attacks~\cite{Goodfellow2015ExplainingAH, kurakin2016adversarial, madry2017towards}. These approaches generate adversarial samples based on the gradient of the loss function with respect to input images.
Another flavor of attacks is based on solving optimization problems to generate adversarial samples~\cite{szegedy2013intriguing, carlini2017towards}. In the black-box setting, only benign images and their class labels are given, meaning attackers can only query the target model. 
Black-box attacks mainly leverage the free query and adversarial transferability to train substitute models~\cite{Hu2017GeneratingAM, DBLP:journals/corr/PapernotMG16, Shi2019CurlsW, 10.1145/3052973.3053009} or directly estimate the target model gradients~\cite{chen2017zoo, cheng2018query, tu2019autozoom} to generate adversarial examples. To avoid the transferability assumption and the overhead of gathering data to train a substitute model, several works proposed local-search-based black-box attacks to generate adversarial samples directly in the input domain~\cite{8014906, li2019nattack, brendel2018decision}.  

In the physical world, adversarial samples are captured by cameras as inputs to target networks, involving camera hardware ISPs and optical systems. A variety of strategies have been developed to guard the effectiveness of the adversarial patterns in the wild~\cite{athalye2018synthesizing, eykholt2018robust, jan2019connecting, duan2020adversarial}. These methods typically assume that the camera acquisition and subsequent hardware processing do not alter the adversarial patterns. However, Phan~\etal~\cite{phan2021adversarial} have recently realized attacks of individual camera types by exploiting slight differences in their hardware ISPs and optical systems. 

\vspace{-4mm}
\subsection{Defense Methods}
\vspace{-2mm}
In response to adversarial attack methods, there have been significant efforts in constructing defenses to counter those attacks. These include adversarial training ~\cite{madry2017towards}, input transformation \cite{JPEG_eval, bahat2019natural}, defensive distillation~\cite{papernot2016distillation}, dynamic models~\cite{wu2020adversarial}, loss modifications~\cite{pang2019rethinking}, model ensemble~\cite{sen2020empir} and robust architecture~\cite{guo2020meets}. Note that with the ongoing intense arms race between attacks and defenses, no defense methods are immunized to all existing attacks~\cite{akhtar2021advances}. We next analyze the two representative categories of defense methods:
\vspace{1mm}

\tb{Adversarial Training (AT):} The idea of AT is the following: in each training loop, it augments training data with adversarial examples generated by different attacks. AT is known to ``overfit'' to the attacks ``seen'' during training and has been demonstrated to be vastly effective in defending those attacks. However, AT does not generalize well on ``unseen'' attacks~\cite{stutz2020confidence}. Furthermore, iteratively generating adversarial images is time-consuming, taking 3-30 times longer than standard training before the model converges~\cite{shafahi2019adversarial}. Multiple methods have been proposed to reduce the training time, making AT on large datasets (\eg, ImageNet) possible~\cite{wong2020fast, wang2019bilateral, Gong_2021_CVPRMaxup, zheng2020efficient}. Even so, for each specific model, it still requires an extra adversarial training process and suffers from cross-domain attacks. Besides the target model, it is also worth noting that adversarial examples can be used to train the input preprocessing models. \cite{GuidedDenoiser}.
\vspace{1mm}

\tb{Input Transformation (IT):} IT, as an image pre-processing approach, aims to remove adversarial patterns to counter attacks. A considerable number of IT methods have been proposed such as JPEG compression~\cite{JPEG_eval, Feature_dist,das2017keeping}, randomization~\cite{xie2017mitigating}, image quilting~\cite{guo2017countering}, pixel deflection~\cite{prakash2018deflecting}, and deep-learning-based approaches~\cite{jia2019comdefend, liu2019feature, samangouei2018defense}. These IT methods can seamlessly work with different downstream models and tasks. More importantly, the IT methods can be easily combined with other model-specific defense methods to offer a stronger defense.  
\vspace{2mm}

Our work falls into the IT category. Unlike the existing IT methods to focus on the preprocessing in the RGB distribution, the proposed approach leverages intermediate natural RAW distribution to remove adversarial patterns, which is the first work to exploit RAW distribution in the adversarial defense domain.

\vspace{-5pt}
\section{Sensor Image Formation}
\label{sec:isp}
\vspace{-2mm}
In this section, we review how a RAW image is formed. In short, when light from the scene enters a camera aperture, it first passes through compound camera optics. Following that is the aperture and shutter, which can be adjusted to define f-number and exposure time. Then the light falls on image sensors (\eg, CCD and CMOS), where the photons are color-filtered and converted into electrons. Finally, the electrons are converted to digital values, comprising a RAW image. We refer the reader to Karaimer and Brown~\cite{karaimer2016software} for a detailed review.
\vspace{2mm}

\begin{figure}[t]
\vspace{-0mm}
\begin{center}
\begin{minipage}{0.55\linewidth}
 \begin{adjustbox}{width=1\linewidth}
\includegraphics[width=1\linewidth, trim=0 0 0 0,clip]{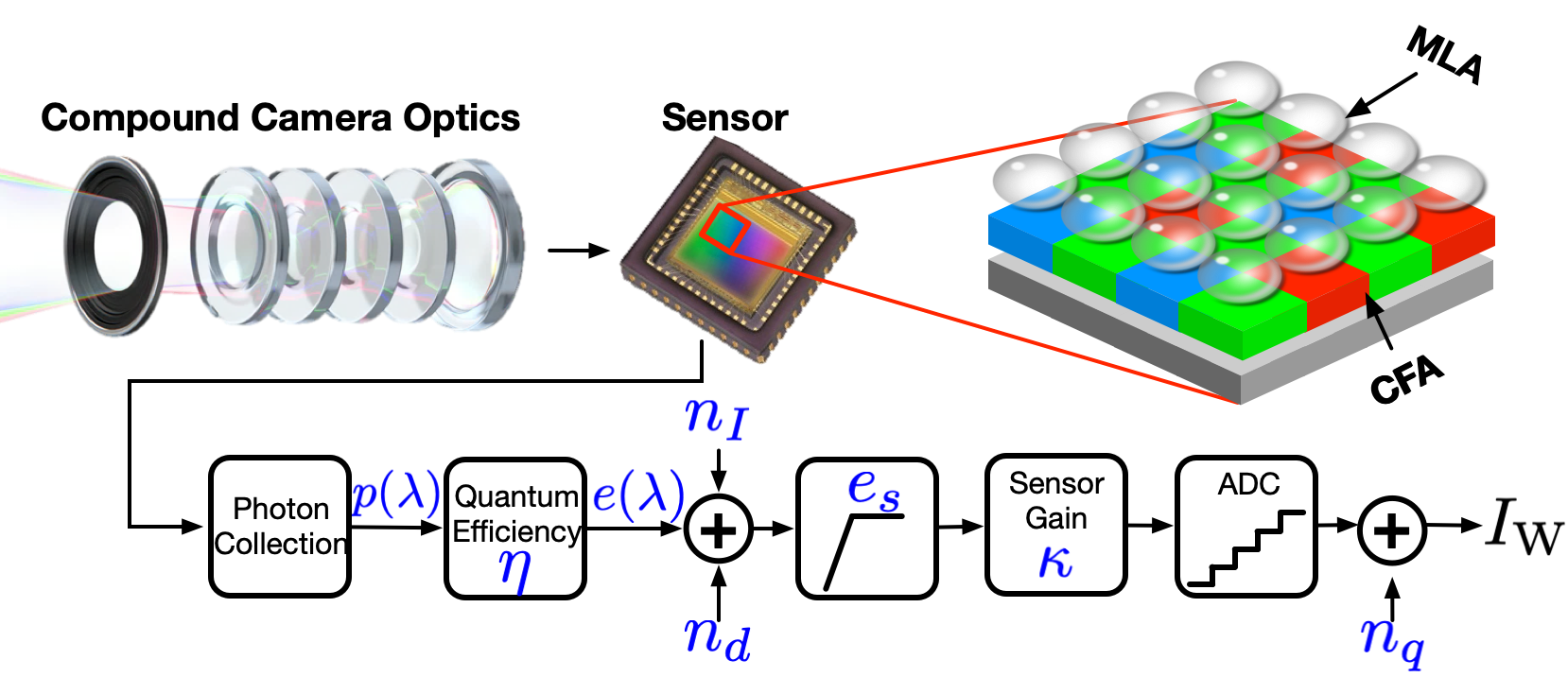}
\end{adjustbox}
\end{minipage}
\hfill
\begin{minipage}{0.44\linewidth}
\captionsetup{font=scriptsize,labelfont=scriptsize}
\caption{
{Overview of the RAW imaging pipeline model. The scene light field is captured by compound camera optics, and then it is gathered by an MLA layer and fed through a CFA layer. The color-filtered photons are converted into electrons based on quantum efficiency before adding dark current and noise. Next, the converted electrons are clipped based on the maximum well capacity, $e_s$, and scaled by a sensor gain factor $\kappa$. Finally, an ADC converts the analog signal into a digital readout with quantization noise $n_q$, $I_\text{W}$. 
}
}
\label{fig:raw_formation}
\end{minipage}
\end{center}
\vspace{-10mm}
\end{figure}

\vspace{-1mm}
\tb{Compound Camera Optics:}
A compound lens consisting of a sequence of optics is designed to correct optical aberrations. When a scene radiance, $I_{\text{SCENE}}$ (in the form of a light field) enters a compound lens, the radiance is modulated by the complex optical pipelines and generates the image $I_{\text{O}}$, that appears on an image sensor surface. Compound optics can be modeled by spatially-varying point spread functions (PSFs)~\cite{Tseng2021DeepCompoundOptics}.

\tb{Color Image Sensor Model:} 
A conventional color image sensor has three layers. On the top is a micro-lens array (MLA) layer; the bottom is a matrix of small potential wells; a color filter array (CFA) layer sits in the middle. When $I_{O}$ falls on a color image sensor, photons first goe through the MLA to improve light collection. Next, light passes through the CFA layer, resulting in a mosaic pattern of the three stimulus RGB colors. 
Finally, the bottom layer collects the color-filtered light and outputs a single channel RAW image, $I_W$. 

The detailed process is illustrated in Figure~\ref{fig:raw_formation}. In particular, at the bottom layer, a potential well counts photons arriving at its location $(x, y)$ and converts the accumulated photons into electrons, and the conversion process is specified by the detector quantum efficiency. 
During the process, electrons could be generated by other resources, called electron noise. 
Two common electron noise types are the dark noise $n_d$, which is independent of light; and dark current $n_I$, which depends on the sensor temperature. These follow normal and Poisson distributions, respectively~\cite{Tseng2021DeepCompoundOptics}.
%
Next, the converted electrons are clipped based on the maximum well capacity, $e_s$, and scaled by a sensor gain factor $\kappa$. Finally, the modulated electrons are converted to digital values by an analog-to-digital converter (ADC), which involves quantization of the input and introduces a small amount of noise, $n_q$. 

Mathematically, a pixel of a RAW image, $I_W$, at position $(x, y)$ can be defined as:
\begin{equation}
    I_W(x,y) = b + n_q + \kappa \min(e_{s}, n_d+n_I+\sum_{\lambda}e(x, y, \lambda)),
\end{equation}
where $b$ is the black level, level of brightness with no light; $e(x,y,\lambda)$ is the number of electrons arrived at a well at position $(x, y)$ for wavelength $\lambda$.

This image formation model reveals that besides the natural scene being captured, RAW images heavily depend on the \emph{specific stochastic natures of the optics, color filtering, sensing, and readout components}. The proposed method exploits these statistics.

\vspace{-5pt}
\section{Raw Image Domain Defense} 


\vspace{-2mm}

In this section, we describe the proposed defense method, which leverages the distribution of RAW measurements as a prior to project adversarially perturbed RGB images to benign ones. Given an adversarial input, existing defense approaches learn an RGB-to-RGB projection from the adversarially perturbed distribution of RGB images, $Y'$, to the closest point in corresponding RGB natural distribution, $Y$. We use the operator $T: Y' \rightarrow Y$ for this projection operation. As this RGB distribution $Y$ empirically sampled from the ISP outputs of diverse existing cameras, it also ingests diverse reconstruction artifacts, making it impossible to exploit photon-flux specific cues, e.g., photon shot noise, optical aberrations, or camera-specific readout characteristics -- as image processing pipelines are designed to remove such RAW cues.


\begin{figure}[t]
\centering
\begin{minipage}{.5\textwidth}
  \centering
  \includegraphics[width=.99\linewidth]{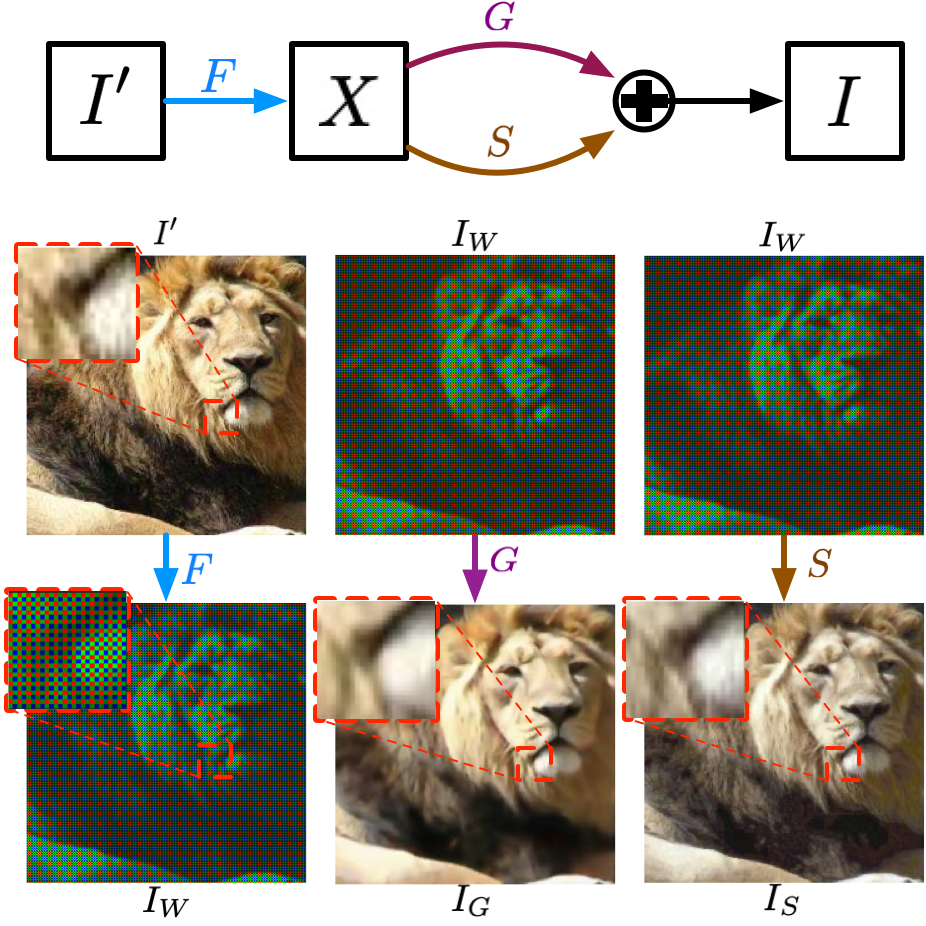}
  \vspace{-5mm}
  \caption{\footnotesize{Overview of the proposed defense approach, see text. We note that the resolution of the RAW image (RGGB) is twice larger than that of the RGB image. We linearly scaled the RAW image in this figure for better visualization.}}
  \label{fig:architecture}
  \vspace{-7mm}
\end{minipage}%
\hfill
\begin{minipage}{.48\textwidth}
  \centering
  \includegraphics[width=.99\linewidth]{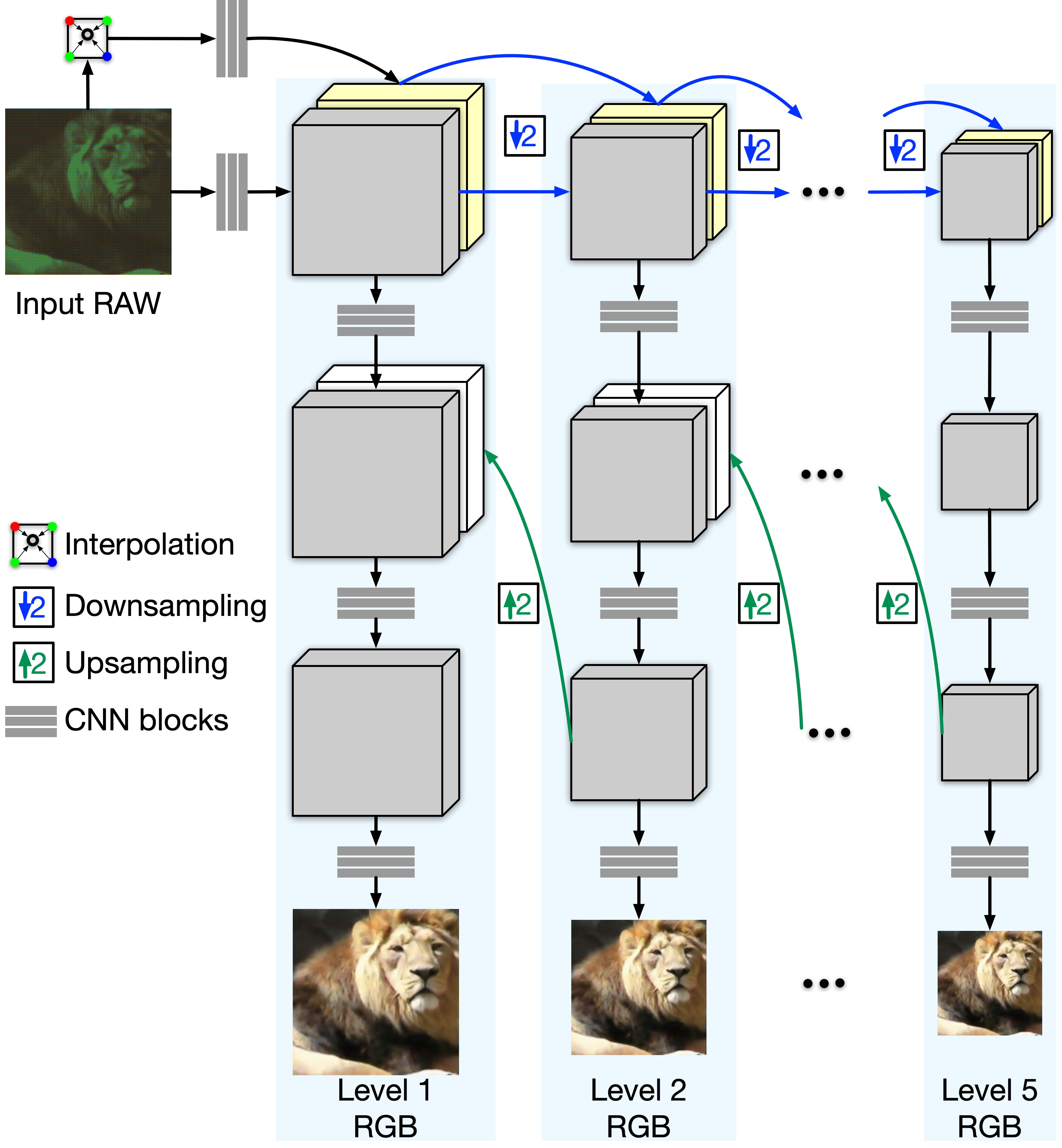}
  \vspace{-5mm}
  \caption{\footnotesize{The architecture of the $G$ operator, which is adopted and modified from PyNet~\cite{ignatov2020replacing}. The finer operator level exploits upsampled coarser-level features to reconstruct the RGB output. The model is trained sequentially in a coarse-to-fine manner. }}
  \label{fig:DeepISP}
  \vspace{-7mm}
\end{minipage}
\end{figure}



Departing from existing methods, as illustrated in Figure~\ref{fig:RawDistribution}, we learn a mapping from $Y'$ to $Y$ via an intermediate RAW distribution, $X$, which incorporates these RAW statistics of natural images, such as sensor photon counts, multispectral color filter array distributions and optical aberrations. To this end, the approach leverages three specially designed operators: $F: Y' \rightarrow X$, $G: X \rightarrow Y$, and $S: X \rightarrow Y$. Specifically, the $F$ operator is a learned model, which maps an adversarial sample from its adversarial distribution to its corresponding RAW sample in the natural image distribution of RAW images. Operator $G$ is another learned network that performs an ISP reconstruction task, \ie, it converts a RAW image to an RGB image. In theory, our goal can be achieved with these two operators by concatenating both $G(F(\cdot)): Y' \rightarrow X \rightarrow Y$. However, as these two operators are differentiable models, the potential adversary may still be able to attack the model if, under stronger attack assumptions,  he has full access to the weight of preprocessing modules.  To address this issue, we add the operator $S$, a conventional ISP, to our approach, which is implemented as a sequence of cascaded software-based sub-modules. In contrast to operator $F$, operator $S$ is non-differentiable. Operators $F$ and $G$ are trained separately without end-to-end fine-tuning. Notably, the proposed defense scheme is \emph{entirely model-agnostic} as it does not require any knowledge of potential adversarial attacks. 


For defending against an attack, as shown in Figure~\ref{fig:architecture}, the proposed approach first uses the $F$ operator to map an input adversarial image, $I'$, to its intermediate RAW measurements, $I_W$. Then, $I_W$ is processed separately by the $G$ and $S$ operators to convert it to two images in the natural RGB distribution, denoted as $I_G$ and $I_S$, respectively. Finally, our method outputs a benign image, $I$, in the natural RGB distribution by combining $I_G$ and $I_S$ in a weighted-sum manner.  Mathematically, the defense process is defined as:
\vspace{-1mm}
\begin{equation}
\label{eq:defense}
    I = \omega G(F(I')) +  (1 - \omega) S(F(I')),
\end{equation}
where $\omega$ is a hyper-parameter for weighting the contributions from the two operators $G$ and $S$. In the following sections, we introduce each operator in detail.

\subsection{$F$ Operator: Image to RAW Mapping}
\vspace{-2mm}
We use a small learned encoder-decoder network as the $F$ operator to map an RGB image to its intermediate RAW measurements. The details of network architecture is shown in supplementary.


We train this module in a supervised manner with two $\mathcal{L}_2$ losses. Both of the $\mathcal{L}_2$ losses are calculated based on ground truth (GT) RAW and estimated RAW images. The only difference between the two losses is the input RGB image used for evaluating a RAW image. One is with the original input RGB image, while the other is generated by adding Gaussian noise to the original input RGB image. In doing so, $F$ is trained with the ability to convert both benign and slightly perturbed RGB images to their RAW distribution. We note that the added Gaussian distribution is \emph{different from the correlated noise generated by various adversarial attacks}. Mathematically, given a benign RGB image, $I$, and its corresponding GT RAW measurements, $GT_{W}$, the loss function is defined as:
\vspace{-2 mm}
\begin{align}
    \mathcal{L}_F &= ||F(I), GT_{W}||_{2} + ||F(I + \alpha \varepsilon ), GT_{W}||_{2}, \\
    \varepsilon &\sim \mathcal{N}(\mu, \sigma) \label{eq:L_F},
\end{align}
where $\varepsilon$ is a Gaussian noise with mean, $\mu$, and standard deviation, $\sigma$; $\alpha$ is a random number in the range between 0 and 1, weighting the amount of noise added. We empirically set the $\mu$ and $\sigma$ to 0 and 1, respectively.


\vspace{-4mm}
\subsection{$G$ Operator: Learned ISP} 
\vspace{-2mm}
The $G$ operator, learned network, converts the $I_w$ generated by the $F$ operator to an RGB image. The challenge of converting RAW images to RGB images is that the process requires both global and local modifications. The global modifications aim to change the high-level properties of the image, such as brightness and white balance. In contrast, the local modifications refer to low-level processing like texture enhancement, sharpening, and noise removal. During the image reconstruction process, an effective local process is expected to be guided by global contextual information, which requires the information exchange between global and local operations. This motivates us to leverage a pyramidal convolutional neural network to fuse global and local features for optimal reconstruction results. We adopt and modify architecture similar to the PyNet~\cite{ignatov2020replacing}. As shown in Figure~\ref{fig:DeepISP}, the network has five levels, the $1st$ level is the finest, and the $5th$ level is the coarsest. The finer-level uses upsampled features from the coarser-level by concatenating them. We modified PyNet by adding an interpolation layer before the input of each level, interpolating the downsampled RAW Bayer pattern. This practice facilitates learning as the network only needs to learn the residuals between interpolated RGB and ground truth RGB, leading to better model performance.


The loss function for this model consists of three components: perceptual, structural similarity, and $\mathcal{L}_2$ loss. The perceptual and $\mathcal{L}_2$ loss functions are adopted to ensure the fidelity of the reconstructed image, and the structural similarity loss function \cite{wang_simoncelli_bovik} is used to enhance the dynamic range. Given an input RAW image, $I_W$, and the corresponding GT RGB image $GT_I$, the loss function can be mathematically defined as: 
\vspace{-3mm}
\begin{align}
	\mathcal{L}_G^i = & \beta^i \mathcal{L}_{Perc}(G(I_W), GT_I) + \gamma^i \mathcal{L}_{SSIM}((G(I_W), GT_I)) \nonumber \\
	&+ \mathcal{L}_2(G(I_W), GT_I) \quad\quad \text{for} \quad i \in [1, 5] \label{eq:D},
\end{align}
where $i$ represents the training level. As the model is trained in a coarse-to-fine manner,  different losses are used for each level $i$. $\mathcal{L}_{Perc}$, $\mathcal{L}_{SSIM}$, and $\mathcal{L}_2$ represents the perceptual loss calculated with VGG architecture, structural similarity loss, and $\mathcal{L}_2$ loss, respectively; $\beta^i$ and $\gamma^i$ are the two weighting hyper-parameters, which are set empirically. The model is trained sequentially in a coarse-to-fine manner, \ie, from $i=5$ to $i=1$. 

\vspace{-4mm}
\subsection{$S$ Operator: Conventional ISP}
\vspace{-2mm}
The $S$ operator has the same functionality as the $G$ operator, converting a RAW image to an RGB image. Unlike the $G$ operator, the $S$ operator offers the functionalities of a conventional hardware ISP pipeline using a sequence of cascaded sub-modules, and it is non-differentiable. 

While we may exploit the ISP pipeline of any digital camera we can extract raw and post-ISP data from, we use a software-based ISP pipeline consisting of the following components: Bayer demosaicing, color balancing, white balancing, contrast improvement, and colorspace conversion sub-modules. Based on the Zurich-Raw-to-RGB dataset~\cite{Ignatov2020ReplacingMC}, we manually tune the hyperparameters of all sub-modules to find the optimal ones that offer the converted RGB image with similar image quality to the original RGB ones. We refer the reader to the Supplementary Material for a detailed description.

\vspace{-4mm}
\subsection{Operator Training}
\vspace{-2mm}
We use the Zurich-Raw-to-RGB dataset \cite{Ignatov2020ReplacingMC} to train the $F$ and $G$ operators. The Zurich-Raw-to-RGB dataset consists of 20,000 RAW-RGB image pairs, captured using a Huawei P20 smartphone with a 12.3 MP Sony Exmor IMX380 sensor and a Canon 5D Mark IV DSLR. 
Both of the $F$ and $G$ operators are trained in PyTorch with Adam optimizer on NVIDIA A100 GPUs. We set the learning rate to $1\mathrm{e}{-4}$ and $5\mathrm{e}{-5}$ for training F and G operators, respectively. The hyperparameters used in our approach have the following settings: $\omega=0.7$ in Eq.~\ref{eq:defense}; $\mu=0$ and $\sigma=1$ for the Gaussian noise used in Eq.~\ref{eq:L_F} ; In Eq.~\ref{eq:D}, $\beta^i$ is set to 1 for $i\in[1,3]$ and $0$ for $i\in[4,5]$; $\gamma^i$ is set to 1 for $i=1$ and $0$ for $i\in[2,5]$. 
\vspace{-2mm}

\vspace{-5pt}
\section{Experiments \& Analysis} 
\begin{table*}[h]
		\vspace{-4mm}
		\begin{minipage}{0.71\linewidth}
			\begin{adjustbox}{width=1.4\linewidth}
				{\small
					\addtolength{\tabcolsep}{0pt}
					\begin{tabular}{l|cccccccccccc}
						\hline
						\hline
						 & \multicolumn{2}{c}{FSGM} & \multicolumn{2}{c}{PGD} & \multicolumn{2}{c}{BIM} & \multicolumn{2}{c}{DeepFool} & \multicolumn{2}{c}{C\&W}  & NewtonFool & BPDA\\
						 & $2/255\uparrow$ & $4/255\uparrow$ &$2/255\uparrow$ & $4/255\uparrow$ & $2/255\uparrow$ & $4/255\uparrow$ & $L_{\infty}\uparrow$ & $L_{2}\uparrow$ & $L_{\infty}\uparrow$ & $L_{2}\uparrow$  &  $L_{\infty}\uparrow$ & $L_{\infty}\uparrow$
						\\
						\hline
						 & \multicolumn{11}{c}{\textbf{ResNet-101}}
						\\[2pt]		
					     JPEG-Defense\cite{JPEG_eval} & 33.14 & 20.71 &	45.19 & 21.74 &	36.78  & 8.5  &	53.16 & 45.69 &	59.06 & 52.01 &	24.65 & 0.08\\ 
						 TVM\cite{guo2017countering}  & 43.75 & 40.02 &	45.46 & 44.35 &	44.86 & 41.93 &	47.69 & 39.89 &	45.51 & 40.44 &	22.6 & 6.39
						 \\ 
						Randomized Resizing \& Padding\cite{xie2017mitigating} & 45.21 & 34.97 &	45.38 &  27.75 &	40.04 & 18.04  &	73.06 &62.47 &	66.53 &  59.87 &	27.93 & 2.66
						   \\ 
						HGD \cite{liao2018defense}& 54.75 & 43.85 &	55.26 & 50.05 &	56.74 &  48.61  &	64.34 & 58.13 &	59.98 & 52.88 &	27.70 &0.03
								\\		
						Pixel-Deflection\cite{prakash2018deflecting} & 54.56 & 35.14 & 60.68 & 34.86 &	58.71 & 41.91  &	\textbf{75.97} & \textbf{64.13} & 66.29 & 60.91 &	28.81 & 1.87
						 \\ 
						ComDefend\cite{jia2019comdefend} & 48.21 & 36.51 &	53.28 & 48.38 &	51.39 &  42.01  &	63.68 & 55.62 &	58.53 & 50.38 &	26.46 & 0.03
								\\				
						Proposed Method   & \textbf{66.02} & \textbf{58.85}  & \textbf{68.34}  & \textbf{66.17} &	\textbf{66.91} & \textbf{63.01} &	72.04 & 63.52 &	\textbf{71.40} & \textbf{67.33}  &	\textbf{40.96} & \textbf{38.85}
						\\
						\hline
						 & \multicolumn{11}{c}{\textbf{InceptionV3}}
						\\[2pt]	
						JPEG-Defense\cite{JPEG_eval} & 31.97 & 20.25 & 43.34 & 21.15 &  34.68 & 8.55& 51.20 & 43.49 & 55.00 & 50.39 & 24.06 & 0.12
						\\ 
						TVM\cite{guo2017countering}  & 42.47 & 37.23 & 42.75 & 41.61 & 42.80 & 39.71 & 45.21  & 37.39  & 43.27 & 37.51 & 23.05 & 4.58

						 \\ 
						Randomized Resizing \& Padding\cite{xie2017mitigating} & 41.86 & 34.49  & 43.41 & 25.60 & 39.42 & 16.62  & 70.24 & 58.65 & 63.24  & 55.62 & 27.55 & 2.09

						   \\ 
						   
                        HGD \cite{liao2018defense}& 52.83 & 40.99  & 50.35 & 47.62 & 56.02 & 47.78  & 60.33 & 56.61 & 59.55  & 52.0 & 26.84 & 0.03 
                        \\
						Pixel-Deflection\cite{prakash2018deflecting} & 51.42  & 34.27  & 56.13   & 32.49  & 56.18  & 39.13   & \textbf{71.16}  & \textbf{61.58}  & 	61.94  & 57.58  & 28.01 & 1.56

						 \\ 
						ComDefend\cite{jia2019comdefend} & 47.00 & 35.34  & 49.99  & 46.15 & 48.74  & 39.58  & 60.01  & 52.47  & 55.85  & 47.70  & 25.44 & 0.03

								\\				
						Proposed Method   & \textbf{63.03}  & 	\textbf{56.34}  & \textbf{65.69}  & 	\textbf{63.03}  & \textbf{64.77}  & 	\textbf{59.49}   & 69.25   & 60.04   & \textbf{66.97}  & \textbf{64.69}  & \textbf{38.01} & \textbf{36.43} 

						  \\ 
						  \hline
						  \hline
						  
					\end{tabular}\\
				}
			\end{adjustbox}
		\end{minipage}
		\vspace{1mm}
		\captionsetup{font=scriptsize,labelfont=scriptsize}
		\caption{{\textbf{Quantitative Comparisons on ImageNet} We evaluate Top-1 Accuracy on ImageNet and compare the proposed method to existing input-transformation methods. The best Top-1 accuracies are marked in bold. Our defense method offers the best performance in all settings, except for the DeepFool attack.}}
		 \label{tbl:imagenet}
		 
	\vspace{-5mm}
\end{table*}

\begin{figure*}[t]
	\centering
	\resizebox{\linewidth}{!}{
	\renewcommand{\arraystretch}{0.5}
	\begin{tabular}{ccccccccccc}
	
		\includegraphics[width=0.35\columnwidth, trim={0cm 0cm 0cm 0cm},clip]{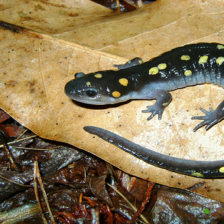}&
		\includegraphics[width=0.35\columnwidth, trim={0cm 0cm 0cm 0cm},clip]{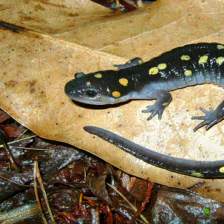}&
		\includegraphics[width=0.35\columnwidth, trim={0cm 0cm 0cm 0cm},clip]{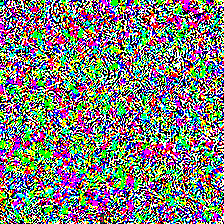}&
		\includegraphics[width=0.35\columnwidth, trim={0cm 0cm 0cm 0cm},clip]{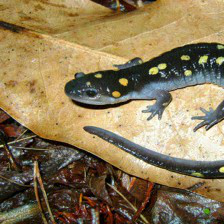} &	
		\includegraphics[width=0.35\columnwidth, trim={0cm 0cm 0cm 0cm},clip]{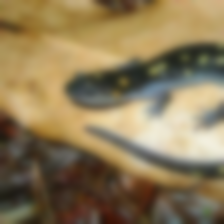} &
		\includegraphics[width=0.35\columnwidth, trim={0cm 0cm 0cm 0cm},clip]{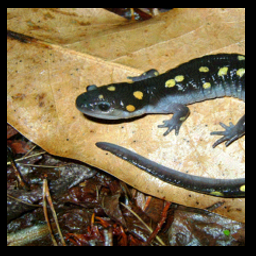} & 
		\includegraphics[width=0.35\columnwidth, trim={0cm 0cm 0cm 0cm},clip]{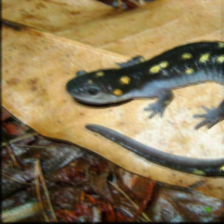} &
		\includegraphics[width=0.35\columnwidth, trim={0cm 0cm 0cm 0cm},clip]{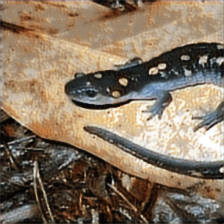} &	
		\includegraphics[width=0.35\columnwidth, trim={0cm 0cm 0cm 0cm},clip]{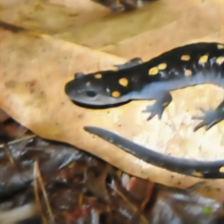}&
		\includegraphics[width=0.35\columnwidth, trim={0cm 0cm 0cm 0cm},clip]{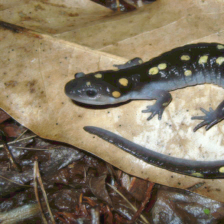} &
		\includegraphics[width=0.35\columnwidth, trim={0cm 0cm 0cm 0cm},clip]{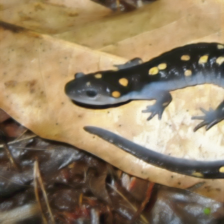} 		
		\\[.2cm]

		\includegraphics[width=0.35\columnwidth, trim={0cm 0cm 0cm 0cm},clip]{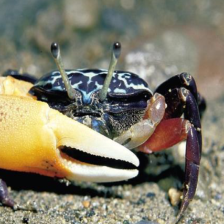}&
		\includegraphics[width=0.35\columnwidth, trim={0cm 0cm 0cm 0cm},clip]{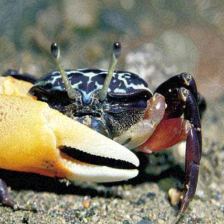}&
		\includegraphics[width=0.35\columnwidth, trim={0cm 0cm 0cm 0cm},clip]{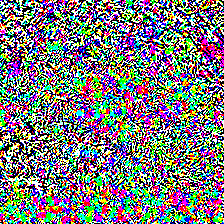}&
		\includegraphics[width=0.35\columnwidth, trim={0cm 0cm 0cm 0cm},clip]{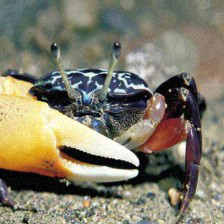} &	
		\includegraphics[width=0.35\columnwidth, trim={0cm 0cm 0cm 0cm},clip]{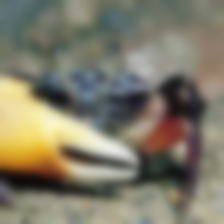} &
		\includegraphics[width=0.35\columnwidth, trim={0cm 0cm 0cm 0cm},clip]{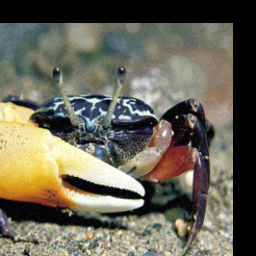} & 
		\includegraphics[width=0.35\columnwidth, trim={0cm 0cm 0cm 0cm},clip]{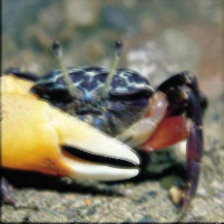} &
		\includegraphics[width=0.35\columnwidth, trim={0cm 0cm 0cm 0cm},clip]{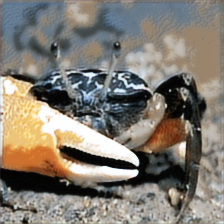} &	
		\includegraphics[width=0.35\columnwidth, trim={0cm 0cm 0cm 0cm},clip]{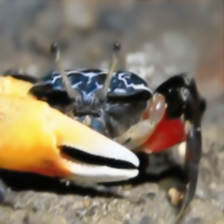}&
		\includegraphics[width=0.35\columnwidth, trim={0cm 0cm 0cm 0cm},clip]{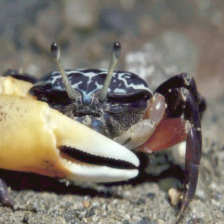} &
		\includegraphics[width=0.35\columnwidth, trim={0cm 0cm 0cm 0cm},clip]{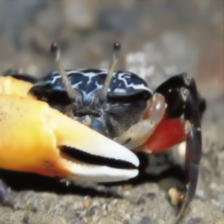} 	
		\\[.2cm]
		
		\includegraphics[width=0.35\columnwidth, trim={0cm 0cm 0cm 0cm},clip]{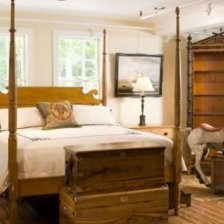}&
		\includegraphics[width=0.35\columnwidth, trim={0cm 0cm 0cm 0cm},clip]{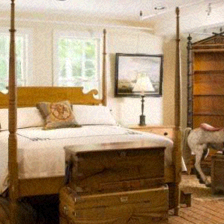}&
		\includegraphics[width=0.35\columnwidth, trim={0cm 0cm 0cm 0cm},clip]{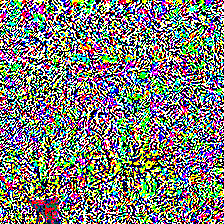}&
		\includegraphics[width=0.35\columnwidth, trim={0cm 0cm 0cm 0cm},clip]{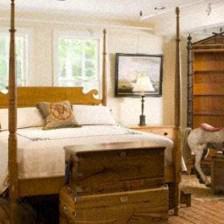} &	
		\includegraphics[width=0.35\columnwidth, trim={0cm 0cm 0cm 0cm},clip]{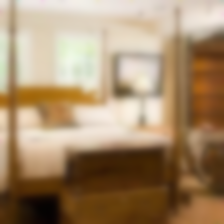} &
		\includegraphics[width=0.35\columnwidth, trim={0cm 0cm 0cm 0cm},clip]{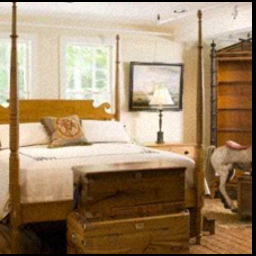} & 
		\includegraphics[width=0.35\columnwidth, trim={0cm 0cm 0cm 0cm},clip]{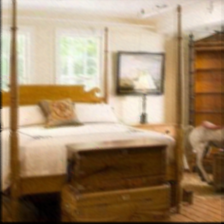} &
		\includegraphics[width=0.35\columnwidth, trim={0cm 0cm 0cm 0cm},clip]{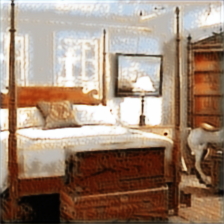} &	
		\includegraphics[width=0.35\columnwidth, trim={0cm 0cm 0cm 0cm},clip]{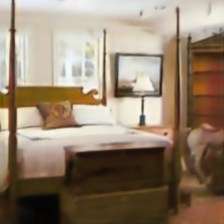}&
		\includegraphics[width=0.35\columnwidth, trim={0cm 0cm 0cm 0cm},clip]{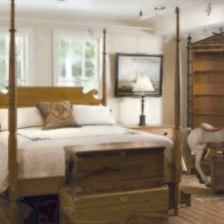} &
		\includegraphics[width=0.35\columnwidth, trim={0cm 0cm 0cm 0cm},clip]{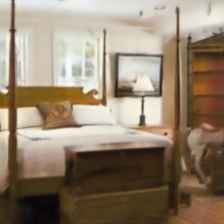} 	
		\\[.2cm]		
		\centering
		{ \huge{Clean Image}}&
		{ \huge{Adversarial Image}}&
		{ \huge{Perturbation}}&		
		{ \huge{JPEG-Defense\cite{JPEG_eval}}} &
		{ \huge{TVM\cite{guo2017countering} }} &
		{ \huge{Resizing \& Padding\cite{xie2017mitigating} }} &
		{ \huge{Pixel-Deflection\cite{prakash2018deflecting}}} &
		{ \huge{ComDefend\cite{jia2019comdefend}}} & 
		{ \huge{Operator $G$}}&
		{ \huge{Operator $S$}} &
		{ \huge{Proposed Method}} 		
		\\		
	\end{tabular} 
	}

	\vspace{-2mm}
	\captionsetup{font=footnotesize,labelfont=footnotesize}
    \caption{Qualitative outputs of the proposed method along with both $G$ and $S$ operators, and state-of-the-art defense methods on the ImageNet dataset, see text.}
    \label{fig:result_reconstruction}
    \vspace{-5mm}
\end{figure*}

\begin{table*}[t]
		 
		 \vspace{-2mm}
		\begin{minipage}{0.71\linewidth}
			\begin{adjustbox}{width=1.4\linewidth}
				{\small
					\addtolength{\tabcolsep}{0pt}
					\begin{tabular}{l|cccccccc}
						\hline
						\hline
						 & \multicolumn{2}{c}{FSGM} & 
						 \multicolumn{2}{c}{PGD} & 
						 \multicolumn{2}{c}{BIM} & 
						 \multicolumn{2}{c}{DAG}  \\
						 & $L_{\infty} = 2/255\uparrow$ & $L_{\infty} = 4/255\uparrow$ &$L_{\infty} = 2/255\uparrow$ & $L_{\infty} = 4/255\uparrow$ & $L_{\infty} = 2/255\uparrow$ & $L_{\infty} = 4/255\uparrow$ & $L_{\infty} = 2/255\uparrow$ & $L_{\infty} = 4/255\uparrow$ \\
						\hline 
						JPEG-Defense\cite{JPEG_eval} & 37.41 & 32.27 & 24.53 &  6.21 & 25.74 & 10.18 & 14.12 & 5.66
						\\ 
						TVM\cite{guo2017countering} \cite{guo2017countering} & 42.64 & 41.53 & 45.55 & 42.24 & 44.51 & 38.44 & 31.88 & 25.56
						 \\ 
				        HGD \cite{liao2018defense}& 43.39 & 40.82 & 44.54 & 40.88 & 40.03 & 39.95 & 28.61 & 22.36
				        \\
						Pixel-Deflection\cite{prakash2018deflecting} & 44.13 & 41.88 & 46.38 & 42.32 & 44.78 & 37.22 & 30.73 & 24.61
						 \\ 
						ComDefend\cite{jia2019comdefend} & 45.57 & 39.23 & 44.85 & 41.14 & 42.71 & 36.12 &  28.94 & 23.36
						\\				
						Proposed Method   & \textbf{52.35} & \textbf{48.04} & \textbf{53.41} & \textbf{49.59} & \textbf{54.86} & \textbf{50.51} & \textbf{40.35} & \textbf{37.88}
						  \\ 
						\hline
						\hline
						
					\end{tabular}\\
				}
			\end{adjustbox}
		\end{minipage}
		\vspace{1mm}
		\captionsetup{font=scriptsize,labelfont=scriptsize}
		\caption{{\textbf{Quantitative Comparison to SOTA Input-Transformation Defense Methods on the COCO dataset.} We evaluate all methods on mean IoU (mIoU) and mark the best mIoU in bold. Our defense method offers the best performance in all settings.}}
		
		 \label{tbl:coco}

	\vspace{-5mm}
\end{table*}

\begin{table*}

		\begin{minipage}{0.71\linewidth}
			\begin{adjustbox}{width=1.4\linewidth}
				{\small
					\addtolength{\tabcolsep}{0pt}
					\begin{tabular}{l|cccccccc}
						\hline
						\hline
						 & \multicolumn{2}{c}{FSGM} & 
						 \multicolumn{2}{c}{PGD} & 
						 \multicolumn{2}{c}{BIM} & 
						 \multicolumn{2}{c}{DAG}  \\
						 & $L_{\infty} = 2/255\uparrow$ & $L_{\infty} = 4/255\uparrow$ &$L_{\infty} = 2/255\uparrow$ & $L_{\infty} = 4/255\uparrow$ & $L_{\infty} = 2/255\uparrow$ & $L_{\infty} = 4/255\uparrow$ & $L_{\infty} = 2/255\uparrow$ & $L_{\infty} = 4/255\uparrow$ \\
						\hline 
						JPEG-Defense\cite{JPEG_eval} & 39.02 & 35.88 & 37.96 & 33.51 & 38.85 & 34.69 & 30.72 & 25.07

						\\ 
						TVM\cite{guo2017countering} & 48.11 & 39.66 & 47.1 & 44.38 & 48.94 & 41.76 & 39.20 & 33.18

						 \\ 
				        HGD \cite{liao2018defense}& 50.68 & 40.06 & 51.24 & 45.92 & 46.80 & 39.74 & 41.15 & 37.23
				        \\
						Pixel-Deflection\cite{prakash2018deflecting} & 53.77 & 44.82 & 54.45 & 47.22 & 55.32 &  48.32 & 46.52 & 39.87
						 \\ 
						ComDefend\cite{jia2019comdefend} & 50.18 & 42.93 & 50.46 & 43.08 & 52.32 & 44.2 & 44.68 & 37.22
						\\				
						Proposed Method   & \textbf{61.68} & \textbf{59.37} & \textbf{64.71} & \textbf{60.23} & \textbf{66.52} & \textbf{61.82} & \textbf{57.83} & \textbf{54.12}
						  \\ 
						\hline
						\hline
						
					\end{tabular}\\
				}
			\end{adjustbox}
		\end{minipage}
		 
		\vspace{1mm}
		\captionsetup{font=scriptsize,labelfont=scriptsize}
		\caption{{\textbf{Quantitative Comparison to SOTA Input-Transformation Defenses on the Pascal VOC dataset}. We evaluate all compared methods for mean average precision (mAP) on Pascal VOC dataset. The best mAP are marked in bold. Our defense method offers the best performance in all settings.}}
		\label{tbl:pascal}
		\vspace{-8mm}
\end{table*}

\vspace{-2mm}
The proposed method acts as an off-the-shelf input preprocessing module, and it requires no additional training to be transferred to different tasks. To validate the effectiveness and generalization capabilities of the proposed defense approach, we evaluate the method on three different vision tasks, \ie, classification, semantic segmentation, and 2D object detection, with corresponding adversarial attacks. 

\vspace{-4mm}
\subsection{Experimental Setup}
\vspace{-2mm}
\tb{Adversarial Attack Methods:} We evaluate our method by defending against the following attacks: FGSM \cite{goodfellow2015explaining}, BIM \cite{kurakin2017adversarial}, PGD \cite{madry2019deep} , C\&W \cite{carlini2017evaluating}, NewtonFool \cite{jang2017objective}, and DeepFool~\cite{moosavidezfooli2016deepfool}. For classification, we use the widely used Foolbox benchmarking suite \cite{rauber2018foolbox} to implement these attack methods. Since Foolbox does not directly support semantic segmentation and object detection, we use the lightweight TorchAttacks library~\cite{kim2021torchattacks} for generating adversarial examples with FGSM, PGD, and BIM attacks. We also evaluate against the DAG \cite{xie2017adversarial} attack, a dedicated attack approach for semantic segmentation and object detection tasks. Moreover, we further evaluate against BPDA \cite{athalye2018obfuscated}, an attack method specifically designed for circumventing input transformation defenses that rely on obfuscated gradients. Applying our method for defending against BPDA, however, requires a slight modification at inference time, see Supplementary Document for details. We note that all applied attacks are untargeted. Definitions of all attack methods are provided in the Supplementary Material.
\vspace{1mm}

\noindent \tb{Baseline Defense Approaches:} We compare to the following input transformation defense methods: JPEG compression~\cite{JPEG_eval}, randomized resizing \& padding~\cite{xie2017mitigating}, image quilting~\cite{guo2017countering}, TVM~\cite{guo2017countering}, HGD \cite{liao2018defense}, pixel deflection~\cite{prakash2018deflecting}, and Comdefend~\cite{jia2019comdefend}. We evaluate all baseline methods on the three vision tasks, except that the randomized resizing \& padding method is omitted in semantic segmentation and object detection tasks as it destroys the semantic structure. We directly adopt the open-source PyTorch implementation for all baseline methods. We use the same training dataset as the one used to train our method for those methods that required training. It is worth noting that all baseline methods do not require adversarial examples for training. 

\vspace{1mm}
\noindent \tb{Evaluation Dataset and Metrics:} For classification, we use the ImageNet validation set and evaluate the Top-1 classification accuracy of all competing defense approaches. For semantic segmentation and object detection, we evaluate on the MS COCO \cite{lin2015microsoft} and Pascal VOC \cite{everingham2010pascal} datasets. The effectiveness of all methods for segmentation and detection is measured by mean Intersection over Union (mIoU) and mean Average Precision (mAP), respectively.  

\vspace{-3mm}
\subsection{Assessment}
\vspace{-1mm}
\tb{Classification:}
We apply a given attack method with ResNet101 and InceptionV3 to generate adversarial samples. For FGSM, BIM, and PGD, we set two different maximum perturbation levels in $L_\infty$ distance, namely $2/255$ and $4/255$. The maximum number of iterations is set to $100$ for both BIM and PGD. For C\&W, NewtonFool, and DeepFool attacks, we generate both $L_\infty$ distance based attacks and $L_2$ distance-based attacks; we choose $100$ update steps for C\&W and NewtonFool, and $50$ for DeepFool; DeepFool requires the number of candidate classes, which is set to $10$ in our experiments. 

The Top-1 classification accuracies of all methods are reported in Table~\ref{tbl:imagenet}. Our approach outperforms the baseline methods with a large margin under all experimental settings except those with DeepFool attacks. Notably, under DeepFool attacks, the differences between the best performer pixel-deflection and ours are marginal.
Moreover, with PGD and BIM attacks, our defense method offers the lowest relative performance degradation when the more vigorous attack is performed (\ie, maximum perturbation increases from $2/255$ to $4/255$). 
Figure~\ref{fig:result_reconstruction} qualitatively underlines the motivation of combining $G$ and $S$ operators in a weighted sum manner. The $G$ operator learns to mitigate the adversarial pattern, i.e., it recovers a latent image in the presence of severe measurement uncertainty, while the $S$ operator can faithfully reconstruct high-frequency details. 
Note that our method is able to generalize well to images from the ImageNet dataset, which typically depict single objects, although it is trained on the Zurich-Raw-to-RGB dataset, consisting of street scenes.

\vspace{1mm}
\noindent \tb{Semantic Segmentation:}
In this task, we conduct experiments with two different types of attacks: the commonly used adversarial attacks, and the attack specially designed for attacking semantic segmentation models. For the former, FGSM, BIM, and PGD are used; We use DAG~\cite{xie2017adversarial}, a dedicated semantic segmentation attack, for the latter. All attacks are based on a COCO-pretrained DeepLabV3 model~\cite{chen2017rethinking}.
%
%
Two different maximum perturbation levels in $L_\infty$ are used (\ie, $2/255$ and $4/255$). The corresponding experimental results are reported in Table~\ref{tbl:coco}. The proposed approach significantly outperforms baseline methods under all experimental settings. Note that no additional training is required to apply the proposed Raw-Defense approach to defend other vision tasks, validating the generalization capabilities of the method.
\vspace{1mm}

\noindent \tb{2D Object Detection:}
The experimental settings are the same as the ones used for semantic segmentation experiments, except that we use a pretrained Faster R-CNN \cite{ren2016faster}. We report the mAPs on the Pascal VOC dataset under different experimental settings in Table \ref{tbl:pascal}. The proposed defense method offers the best defense performance in all experimental settings, indicating that our approach generalizes well to unseen tasks.   

\vspace{-4mm}
\subsection{RAW Distribution Analysis}
\vspace{-2mm}

In this section, we provide additional analysis on the function of the RAW distribution as an intermediate mapping space. Fundamentally, we share the motivation from existing work that successfully exploits RAW data for imaging and vision tasks, including~\cite{steven:dirtypixels2021,tseng2019hyperparameter}. RGB images are generated by processing RAW sensor measurements (see Sec.~\ref{sec:isp}) with an image processing pipeline. This process removes statistical information embedded in the sensor measurements by aberrations in the optics, readout noise, color filtering, exposure, and scene illumination. While existing work directly uses RAW inputs to preserve this information, we exploit it in the form of an \emph{empirical intermediate image distribution}. Specifically, we devise a mapping via RAW space, thereby using RAW data to train network mapping modules, which we validate further below. As a result, we allow the method to remove adversarial patterns not only by relying on RGB image priors but also RAW image priors. We \emph{validate the role of RAW data} in our method in Table~\ref{tbl:ablation_dataset}, resulting in a large Top-1 accuracy drop (\ie, more than $12\%$), when swapping the real RAW distribution to a synthesized one. This is further corroborated in Table~\ref{tbl:ablation_isp}, where the defense breaks down from $71\%$ to $53\%$, when gradually moving from RAW to RGB as intermediate image space. These experiment validate that, the ``rawer" the intermediate image space is, the better the defense performs.
\vspace{1mm}

\noindent
\tb{Effect of Intermediate Mapping Space:} We use the RAW image distribution as the intermediate mapping space in our method. To validate the effectiveness of this choice, we map to other intermediate stages in the processing pipeline, such as demosaicing stage, color balance stage, and the white balance stage. Specifically, we assess how using different stage values as intermediate mapping space affects the defense performance (\ie, we ablate on the intermediate mapping space used). As reported in Table~\ref{tbl:ablation_isp}, we observe that the defense performance gradually decreases as we map via a less RAW intermediate space. In other words, \emph{the ``rawer'' the intermediate image space is, the better performance can be achieved.} This validates the importance and benefit of exploiting RAW distribution in the defense.
\vspace{1mm}

\noindent
\tb{Real RAW Versus Synthetic RAW:} We further ablate on the dataset used to train our model. Specifically, we trained $F$ and $G$  operator on the Zurich-Raw-to-RGB dataset, HDR-RAW-RGB~\cite{hasinoff2016burst} and MIT-RAW-RGB~\cite{gharbi2016deep} respectively and assess how the defense performance changes. Similar to Zurich-Raw-to-RGB dataset, the RAW images in HDR-RAW-RGB are captured by a real camera; however, the ones offered by MIT-RAW-RGB are purely synthesized by reformatting downsampled RGB images into Bayer patterns with handcrafted Gaussian noise. We note that, as such, the MIT-RAW-RGB dataset does not include the RAW distribution cues.
The experimental results are reported in Table~\ref{tbl:ablation_dataset}. As observed, both RAW distribution Zurich-Raw-to-RGB and HDR-RAW-RGB allow us to learn effective adversarial defenses, while a sharp performance degradation occurs when shifting from real RAW distribution to the synthesized one due to the lack of natural RAW distribution cues. This again validates the effectiveness of statistical information in the real RAW distribution when defending against adversarial attacks.
\vspace{-5mm}
\begin{table}[]
\centering
\begin{minipage}{.5\textwidth}
  \centering
			\begin{adjustbox}{width=1\linewidth}
				{\small
					\addtolength{\tabcolsep}{0pt}
					\begin{tabular}{l|ccccc}
						\hline
						\hline
						 & FSGM &  PGD  &  C\&W & NewtonFool & DeepFool   \\
						\hline 
						Zurich-Raw-to-RGB\cite{Ignatov2020ReplacingMC}  & 58.85 & 66.17 & 71.40 & 40.96 & 72.04
						\\ 
						HDR-RAW-RGB\cite{hasinoff2016burst}  & 55.57 & 64.12 & 71.65 & 42.36 & 70.77
						 \\ 
						MIT-RAW-RGB\cite{gharbi2016deep} & 40.52 & 47.29 & 55.13 & 28.52 & 58.49
						  \\ 
						\hline
						\hline
					\end{tabular}
				}
			\end{adjustbox}

  \vspace{1mm}
  \captionsetup{font=scriptsize,labelfont=scriptsize}
  \caption{{\textbf{Quantitative Ablation Study on RAW Training Datasets.} We train $F$ and $G$ two operators with three different RAW-RGB datasets and report the Top-1  defense accuracy on the ImageNet dataset. The RAW images in the Zurich-Raw-to-RGB and HDR-RAW-RGB are captured by real cameras, while the ones in MIT-RAW-RGB are synthesized. We see a sharp performance drop when swapping the real RAW training data to synthetic data due to the lack of natural RAW distribution cues. }}
  \label{tbl:ablation_dataset}
\end{minipage}%
\hfill
\begin{minipage}{.48\textwidth}
  \centering
			\begin{adjustbox}{width=1\linewidth}
				{\small
					\addtolength{\tabcolsep}{0pt}
					\begin{tabular}{l|ccccc}
						\hline
						\hline
						 & FSGM &  PGD  &  C\&W & NewtonFool & DeepFool  \\
						\hline 
						Raw Capture & 57.33 & 65.02 & 70.86 & 40.65 & 70.23
						\\ 
						Demosaic Stage  & 52.93 & 59.83 & 63.29 & 36.76 & 64.81
						 \\ 
						Color Balance Stage & 48.38 & 55.41 & 57.92 & 33.92 & 59.37
						   \\ 
						White Blance Stage & 47.35 &  54.02 &  56.23 &  33.01 & 57.08
						 \\ 
						contrast Improvement Stage & 45.2 &  52.18 &  54.18 &  31.84 & 55.64
						\\				
						Agamma adjustment Stage & 44.4 & 50.91 & 53.08 & 30.57 & 54.19
						  \\ 
						\hline
						\hline
					\end{tabular}
				}
			\end{adjustbox}
  \vspace{1mm}
  \captionsetup{font=scriptsize,labelfont=scriptsize}
  \caption{{\textbf{Effect of Different Intermediate Mapping Spaces.} We report the Top-1 adversarial defense accuracy on ImageNet dataset when mapping to different intermediate mapping spaces that are the steps of the image processing pipeline. The performance drops as the intermediate image space moves from RAW to the RGB output space. This validates the importance and benefit of exploiting RAW
distribution in the defense.}}
  \label{tbl:ablation_isp}
\end{minipage}
\end{table}

\vspace{-15mm}
\subsection{Robustness to Hyper-parameter and Operator Deviations}

\noindent
\tb{Hyper-parameter $\omega$:} We introduced a hyper-parameter $\omega$ for weighting the contributions of the two operators $G$ and $S$. Next, we evaluate how varying values of $\omega$ affect the overall defense accuracy. As reported in Tab.~\ref{tbl:robust3}, we find that, while each attack has a different optimal value of $\omega$, the range 0.6-0.8 provides a good trade-off, and we use 0.7 in our experiments. Limited by space, we report here a subset of attacks.

\begin{table}[h]
		\vspace{-5mm}
		\captionsetup{font=scriptsize}
		\begin{minipage}{0.71\linewidth}
			\begin{adjustbox}{width=1.4\linewidth}
				{\small
					\addtolength{\tabcolsep}{0pt}
					\begin{tabular}{l|ccccccccccc}
						\hline
						\hline
						Hyper-parameter $\omega$ = & 0 & 0.1 & 0.2 & 0.3 & 0.4 & 0.5 & 0.6 & 0.7 & 0.8 & 0.9 & 1 \\
						\hline
						Against FSGM Attack & 64.25 & 64.41 & 64.83 & 65.27 & 65.58 & 65.87 & 65.93 & \textbf{66.02} & 65.75 & 65.53 & 65.39
						 \\ 
						Against C\&W Attack & 69.16 & 69.44 & 69.93 & 70.26 & 70.81 & 70.96 & 71.35 & 71.40 & \textbf{72.70} & 71.28 & 71.07
						 \\ 
						Against DeepFool Attack & 69.55 & 69.84 & 71.19 & 71.51 & 71.88 & 72.35 & \textbf{72.63} & 72.04 & 71.75 & 71.69 & 71.04
						\\
						  \hline
						  \hline
					\end{tabular}\\
				}
			\end{adjustbox}
		\end{minipage}
		\vspace{2mm}
		 \large{ \caption{
		 \textbf{Effect of hyper-parameter $\omega$.} We evaluate the impact of the method hyper-parameter $\omega$ on the effectiveness of the proposed defense method. 
		  \label{tbl:robust3}
		 } }
	\vspace{-15mm}
\end{table}

\begin{table}[]
\centering
\begin{minipage}{.5\textwidth}
  \centering
			\begin{adjustbox}{width=1\linewidth}
            {\small
            \begin{tabular}{l|cccccc}
            	\hline
            	\hline
            	 & $F$-300  & $F$-320 & $F$-340 & $F$-360 & $F$-380 & $F$-400\\
            	\hline
            	$G$-300 & 66.02 & 66.08 & 65.93 & 66.05 & 66.03 & 66.05
            	 \\ 
            	$G$-330 & 66.11 & 66.04 & 65.97 &  66.01 & 65.99 & 66.04
            	 \\ 
            	$G$-360 & 65.98 & 66.04 & 66.02 & 65.89 & 66.08 & 69.01
            	 \\ 
            	$G$-390 & 65.98 & 65.95 & 66.00 & 66.04 & 65.99 &65.94
            	 \\ 
            	  \hline
            	  \hline
            \end{tabular}
            }
			\end{adjustbox}

  \vspace{1mm}
  \captionsetup{font=scriptsize,labelfont=scriptsize}
  \caption{{\textbf{Robustness to Deviations of $F$ and $G$.} We evaluate the defense accuracy when mixing operator from different training epochs.}}
  \label{tbl:robust1}
  \vspace{-11mm}
\end{minipage}%
\hfill
\begin{minipage}{.48\textwidth}
  \centering
			\begin{adjustbox}{width=1\linewidth}
				{\small
					\addtolength{\tabcolsep}{0pt}
					\begin{tabular}{l|cccccc}
						\hline
						\hline
						Gausian Noise $\sigma$ &  0 (no noise) & 0.01 & 0.05 & 0.1 & 0.3 & 0.5 \\
						\hline
						Against FSGM Attack & 66.02 & 66.01 & 65.98 & 65.90 & 65.73 & 65.64
						 \\ 
						Against PGD Attack & 68.34 & 68.30 & 68.22 & 68.10 & 68.03 & 67.83
						 \\ 
						  \hline
						  \hline
					\end{tabular}
				}
			\end{adjustbox}
  \vspace{1mm}
  \captionsetup{font=scriptsize,labelfont=scriptsize}
  \caption{{\textbf{Robustness to Deviations of $F$.} We perturb the output of operator $F$ with Gaussian noise of different standard deviations and report the defense accuracy.}}
  \label{tbl:robust2}
  \vspace{-5mm}
\end{minipage}
\end{table}

\noindent \tb{Deviations of Operators $F$ and $G$:} The operator $F$ and $G$ are trained separately and used jointly at the inference time. We evaluate how deviations in each operator affect the overall performance in two experiments. First, we mix the operators $F$ and $G$ from different training checkpoints and evaluate the effect on the defense accuracy. Tab.~\ref{tbl:robust1} reports that the checkpoint combinations do not result in a failure but only slight deviations of the defense performance. Second, we add varying levels of Gaussian noise $G(0,\sigma)$ to the output of operator $F$ and evaluate how such deviation affects the following steps and the overall defense accuracy. Tab.~\ref{tbl:robust2} reports that such perturbations are not amplified in the following steps, and the defense accuracy only fluctuates slightly. The experiments show that the ISP operators $G$ and $S$ themselves are robust to slight deviation in each component.

\vspace{-5pt}
\vspace{-2mm}
\section{Conclusion}
\vspace{-2mm}
We exploit RAW image data as an empirical latent space in the formulation of the proposed adversarial defense method. Departing from existing defense methods that aim to directly map an adversarially perturbed image to the closest benign image, we exploit large-scale natural image datasets as an empirical prior for sensor-captured images -- before they end up in existing datasets after their transformation through conventional image processing pipelines. This empirical prior allows us to rely on low-level image processing pipelines to design the mappings between the benign and perturbed image distributions. We validate the effectiveness of the method, which is entirely model-agnostic, requires no adversarial examples to train, and acts as an off-the-shelf preprocessing module that can be transferred to diverse tasks. We also provided insight into the working principles of the approach and assess that the method significantly outperforms the comparable baselines. In the future, we plan to explore RAW natural image statistics as an unsupervised prior for image reconstruction and generative neural rendering tasks.

\clearpage
%
%
\bibliographystyle{splncs04}
\bibliography{egbib.bib}

\begin{thebibliography}{10}
\providecommand{\url}[1]{\texttt{#1}}
\providecommand{\urlprefix}{URL }
\providecommand{\doi}[1]{https://doi.org/#1}

\bibitem{akhtar2021advances}
Akhtar, N., Mian, A., Kardan, N., Shah, M.: Advances in adversarial attacks and
  defenses in computer vision: A survey (2021)

\bibitem{athalye2018obfuscated}
Athalye, A., Carlini, N., Wagner, D.: Obfuscated gradients give a false sense
  of security: Circumventing defenses to adversarial examples. In:
  International conference on machine learning. pp. 274--283. PMLR (2018)

\bibitem{athalye2018synthesizing}
Athalye, A., Engstrom, L., Ilyas, A., Kwok, K.: Synthesizing robust adversarial
  examples. In: International conference on machine learning. pp. 284--293.
  PMLR (2018)

\bibitem{bahat2019natural}
Bahat, Y., Irani, M., Shakhnarovich, G.: Natural and adversarial error
  detection using invariance to image transformations. arXiv preprint
  arXiv:1902.00236  (2019)

\bibitem{borkar2020defending}
Borkar, T., Heide, F., Karam, L.: Defending against universal attacks through
  selective feature regeneration. In: Proceedings of the IEEE/CVF Conference on
  Computer Vision and Pattern Recognition. pp. 709--719 (2020)

\bibitem{brendel2018decision}
Brendel, W., Rauber, J., Bethge, M.: Decision-based adversarial attacks:
  Reliable attacks against black-box machine learning models. In: International
  Conference on Learning Representations (2018)

\bibitem{carlini2017towards}
Carlini, N., Wagner, D.: Towards evaluating the robustness of neural networks.
  In: IEEE Symposium on Security and Privacy (2017)

\bibitem{carlini2017evaluating}
Carlini, N., Wagner, D.: Towards evaluating the robustness of neural networks
  (2017)

\bibitem{DBLP:conf/iccv/ChenCDK19}
Chen, C., Chen, Q., Do, M.N., Koltun, V.: Seeing motion in the dark. In: 2019
  {IEEE/CVF} International Conference on Computer Vision, {ICCV} 2019, Seoul,
  Korea (South), October 27 - November 2, 2019. pp. 3184--3193. {IEEE} (2019).
  \doi{10.1109/ICCV.2019.00328}, \url{https://doi.org/10.1109/ICCV.2019.00328}

\bibitem{DBLP:conf/cvpr/ChenCXK18}
Chen, C., Chen, Q., Xu, J., Koltun, V.: Learning to see in the dark. In: 2018
  {IEEE} Conference on Computer Vision and Pattern Recognition, {CVPR} 2018,
  Salt Lake City, UT, USA, June 18-22, 2018. pp. 3291--3300. Computer Vision
  Foundation / {IEEE} Computer Society (2018). \doi{10.1109/CVPR.2018.00347},
  \url{http://openaccess.thecvf.com/content\_cvpr\_2018/html/Chen\_Learning\_to\_See\_CVPR\_2018\_paper.html}

\bibitem{chen2017deeplab}
Chen, L.C., Papandreou, G., Kokkinos, I., Murphy, K., Yuille, A.L.: Deeplab:
  Semantic image segmentation with deep convolutional nets, atrous convolution,
  and fully connected crfs. IEEE transactions on pattern analysis and machine
  intelligence  \textbf{40}(4),  834--848 (2017)

\bibitem{chen2017rethinking}
Chen, L.C., Papandreou, G., Schroff, F., Adam, H.: Rethinking atrous
  convolution for semantic image segmentation (2017)

\bibitem{chen2017zoo}
Chen, P.Y., Zhang, H., Sharma, Y., Yi, J., Hsieh, C.J.: Zoo: Zeroth order
  optimization based black-box attacks to deep neural networks without training
  substitute models. In: Proceedings of the 10th ACM Workshop on Artificial
  Intelligence and Security. pp. 15--26 (2017)

\bibitem{cheng2018query}
Cheng, M., Le, T., Chen, P.Y., Yi, J., Zhang, H., Hsieh, C.J.: Query-efficient
  hard-label black-box attack: An optimization-based approach. arXiv preprint
  arXiv:1807.04457  (2018)

\bibitem{Dai2020AWNetAW}
Dai, L., Liu, X., Li, C., Chen, J.: Awnet: Attentive wavelet network for image
  isp. In: ECCV Workshops (2020)

\bibitem{das2017keeping}
Das, N., Shanbhogue, M., Chen, S.T., Hohman, F., Chen, L., Kounavis, M.E.,
  Chau, D.H.: Keeping the bad guys out: Protecting and vaccinating deep
  learning with jpeg compression. arXiv preprint arXiv:1705.02900  (2017)

\bibitem{steven:dirtypixels2021}
Diamond, S., Sitzmann, V., Julca-Aguilar, F., Boyd, S., Wetzstein, G., Heide,
  F.: Dirty pixels: Towards end-to-end image processing and perception. ACM
  Transactions on Graphics (SIGGRAPH)  (2021)

\bibitem{duan2020adversarial}
Duan, R., Ma, X., Wang, Y., Bailey, J., Qin, A.K., Yang, Y.: Adversarial
  camouflage: Hiding physical-world attacks with natural styles. In:
  Proceedings of the IEEE/CVF Conference on Computer Vision and Pattern
  Recognition. pp. 1000--1008 (2020)

\bibitem{jpeg2016}
Dziugaite, G.K., Ghahramani, Z., Roy, D.M.: A study of the effect of jpg
  compression on adversarial images (2016)

\bibitem{JPEG_eval}
Dziugaite, G.K., Ghahramani, Z., Roy, D.M.: A study of the effect of {JPG}
  compression on adversarial images. CoRR  \textbf{abs/1608.00853} (2016)

\bibitem{everingham2010pascal}
Everingham, M., Van~Gool, L., Williams, C.K., Winn, J., Zisserman, A.: The
  pascal visual object classes (voc) challenge. International journal of
  computer vision  \textbf{88}(2),  303--338 (2010)

\bibitem{eykholt2018robust}
Eykholt, K., Evtimov, I., Fernandes, E., Li, B., Rahmati, A., Xiao, C.,
  Prakash, A., Kohno, T., Song, D.: Robust physical-world attacks on deep
  learning visual classification. In: Proceedings of the IEEE Conference on
  Computer Vision and Pattern Recognition. pp. 1625--1634 (2018)

\bibitem{gharbi2016deep}
Gharbi, M., Chaurasia, G., Paris, S., Durand, F.: Deep joint demosaicking and
  denoising. ACM Transactions on Graphics (TOG)  \textbf{35}(6), ~191 (2016)

\bibitem{Gong_2021_CVPRMaxup}
Gong, C., Ren, T., Ye, M., Liu, Q.: Maxup: Lightweight adversarial training
  with data augmentation improves neural network training. In: Proceedings of
  the IEEE/CVF Conference on Computer Vision and Pattern Recognition (CVPR).
  pp. 2474--2483 (June 2021)

\bibitem{Goodfellow2015ExplainingAH}
Goodfellow, I.J., Shlens, J., Szegedy, C.: Explaining and harnessing
  adversarial examples. CoRR  \textbf{abs/1412.6572} (2015)

\bibitem{goodfellow2015explaining}
Goodfellow, I.J., Shlens, J., Szegedy, C.: Explaining and harnessing
  adversarial examples (2015)

\bibitem{guo2017countering}
Guo, C., Rana, M., Cisse, M., Van Der~Maaten, L.: Countering adversarial images
  using input transformations. ICLR  (2018)

\bibitem{guo2020meets}
Guo, M., Yang, Y., Xu, R., Liu, Z., Lin, D.: When nas meets robustness: In
  search of robust architectures against adversarial attacks. In: Proceedings
  of the IEEE/CVF Conference on Computer Vision and Pattern Recognition. pp.
  631--640 (2020)

\bibitem{hasinoff2016burst}
Hasinoff, S.W., Sharlet, D., Geiss, R., Adams, A., Barron, J.T., Kainz, F.,
  Chen, J., Levoy, M.: Burst photography for high dynamic range and low-light
  imaging on mobile cameras. ACM Transactions on Graphics (ToG)
  \textbf{35}(6),  1--12 (2016)

\bibitem{he2017mask}
He, K., Gkioxari, G., Doll{\'a}r, P., Girshick, R.: Mask r-cnn. In: Proceedings
  of the IEEE international conference on computer vision. pp. 2961--2969
  (2017)

\bibitem{he2016deep}
He, K., Zhang, X., Ren, S., Sun, J.: Deep residual learning for image
  recognition. In: Proceedings of the IEEE conference on computer vision and
  pattern recognition. pp. 770--778 (2016)

\bibitem{Hu2017GeneratingAM}
Hu, W., Tan, Y.: Generating adversarial malware examples for black-box attacks
  based on gan. ArXiv  \textbf{abs/1702.05983} (2017)

\bibitem{ignatov2020replacing}
Ignatov, A., Gool, L.V., Timofte, R.: Replacing mobile camera isp with a single
  deep learning model (2020)

\bibitem{Ignatov2020ReplacingMC}
Ignatov, A.D., Gool, L.V., Timofte, R.: Replacing mobile camera isp with a
  single deep learning model. 2020 IEEE/CVF Conference on Computer Vision and
  Pattern Recognition Workshops (CVPRW) pp. 2275--2285 (2020)

\bibitem{Isola_2017_CVPR}
Isola, P., Zhu, J.Y., Zhou, T., Efros, A.A.: Image-to-image translation with
  conditional adversarial networks. In: Proceedings of the IEEE Conference on
  Computer Vision and Pattern Recognition (CVPR) (July 2017)

\bibitem{jan2019connecting}
Jan, S.T., Messou, J., Lin, Y.C., Huang, J.B., Wang, G.: Connecting the digital
  and physical world: Improving the robustness of adversarial attacks. In:
  Proceedings of the AAAI Conference on Artificial Intelligence. vol.~33, pp.
  962--969 (2019)

\bibitem{jang2017objective}
Jang, U., Wu, X., Jha, S.: Objective metrics and gradient descent algorithms
  for adversarial examples in machine learning. In: Proceedings of the 33rd
  Annual Computer Security Applications Conference. pp. 262--277 (2017)

\bibitem{jia2019comdefend}
Jia, X., Wei, X., Cao, X., Foroosh, H.: Comdefend: An efficient image
  compression model to defend adversarial examples. In: Proceedings of the
  IEEE/CVF Conference on Computer Vision and Pattern Recognition. pp.
  6084--6092 (2019)

\bibitem{Karaimer2016ASP}
Karaimer, H.C., Brown, M.S.: A software platform for manipulating the camera
  imaging pipeline. In: ECCV (2016)

\bibitem{karaimer2016software}
Karaimer, H.C., Brown, M.S.: A software platform for manipulating the camera
  imaging pipeline. In: European Conference on Computer Vision. pp. 429--444.
  Springer (2016)

\bibitem{kim2021torchattacks}
Kim, H.: Torchattacks: A pytorch repository for adversarial attacks (2021)

\bibitem{kurakin2016adversarial}
Kurakin, A., Goodfellow, I., Bengio, S.: Adversarial examples in the physical
  world. arXiv preprint arXiv:1607.02533  (2016)

\bibitem{kurakin2017adversarial}
Kurakin, A., Goodfellow, I., Bengio, S.: Adversarial examples in the physical
  world (2017)

\bibitem{li2019nattack}
Li, Y., Li, L., Wang, L., Zhang, T., Gong, B.: Nattack: Learning the
  distributions of adversarial examples for an improved black-box attack on
  deep neural networks. arXiv preprint arXiv:1905.00441  (2019)

\bibitem{9329084}
Liang, Z., Cai, J., Cao, Z., Zhang, L.: Cameranet: A two-stage framework for
  effective camera isp learning. IEEE Transactions on Image Processing
  \textbf{30},  2248--2262 (2021). \doi{10.1109/TIP.2021.3051486}

\bibitem{GuidedDenoiser}
Liao, F., Liang, M., Dong, Y., Pang, T., Hu, X., Zhu, J.: Defense against
  adversarial attacks using high-level representation guided denoiser. In:
  Proceedings of the IEEE/CVF International Conference on Computer Vision. pp.
  1778--1787 (2018)

\bibitem{liao2018defense}
Liao, F., Liang, M., Dong, Y., Pang, T., Hu, X., Zhu, J.: Defense against
  adversarial attacks using high-level representation guided denoiser. In:
  Proceedings of the IEEE Conference on Computer Vision and Pattern
  Recognition. pp. 1778--1787 (2018)

\bibitem{lin2015microsoft}
Lin, T.Y., Maire, M., Belongie, S., Bourdev, L., Girshick, R., Hays, J.,
  Perona, P., Ramanan, D., Zitnick, C.L., Dollár, P.: Microsoft coco: Common
  objects in context (2015)

\bibitem{Feature_dist}
Liu, Z., Liu, Q., Liu, T., Wang, Y., Wen, W.: Feature {D}istillation:
  {DNN}-oriented {JPEG} compression against adversarial examples. International
  Joint Conference on Artificial Intelligence  (2018)

\bibitem{liu2019feature}
Liu, Z., Liu, Q., Liu, T., Xu, N., Lin, X., Wang, Y., Wen, W.: Feature
  distillation: Dnn-oriented jpeg compression against adversarial examples. In:
  2019 IEEE/CVF Conference on Computer Vision and Pattern Recognition (CVPR).
  pp. 860--868. IEEE (2019)

\bibitem{lu2017safetynet}
Lu, J., Issaranon, T., Forsyth, D.: Safetynet: Detecting and rejecting
  adversarial examples robustly. In: Proceedings of the IEEE International
  Conference on Computer Vision. pp. 446--454 (2017)

\bibitem{madry2017towards}
Madry, A., Makelov, A., Schmidt, L., Tsipras, D., Vladu, A.: Towards deep
  learning models resistant to adversarial attacks. arXiv preprint
  arXiv:1706.06083  (2017)

\bibitem{madry2019deep}
Madry, A., Makelov, A., Schmidt, L., Tsipras, D., Vladu, A.: Towards deep
  learning models resistant to adversarial attacks (2019)

\bibitem{moosavidezfooli2016deepfool}
Moosavi-Dezfooli, S.M., Fawzi, A., Frossard, P.: Deepfool: a simple and
  accurate method to fool deep neural networks (2016)

\bibitem{isp_opt_cvpr20}
Mosleh, A., Sharma, A., Onzon, E., Mannan, F., Robidoux, N., Heide, F.:
  Hardware-in-the-loop end-to-end optimization of camera image processing
  pipelines. In: IEEE Conference on Computer Vision and Pattern Recognition
  (CVPR) (June 2020)

\bibitem{nakkiran2019adversarial}
Nakkiran, P.: Adversarial robustness may be at odds with simplicity. arXiv
  preprint arXiv:1901.00532  (2019)

\bibitem{8014906}
Narodytska, N., Kasiviswanathan, S.: Simple black-box adversarial attacks on
  deep neural networks. In: 2017 IEEE Conference on Computer Vision and Pattern
  Recognition Workshops (CVPRW). pp. 1310--1318 (2017).
  \doi{10.1109/CVPRW.2017.172}

\bibitem{pang2019rethinking}
Pang, T., Xu, K., Dong, Y., Du, C., Chen, N., Zhu, J.: Rethinking softmax
  cross-entropy loss for adversarial robustness. ICLR  (2020)

\bibitem{10.1145/3052973.3053009}
Papernot, N., McDaniel, P., Goodfellow, I., Jha, S., Celik, Z.B., Swami, A.:
  Practical black-box attacks against machine learning. In: Proceedings of the
  2017 ACM on Asia Conference on Computer and Communications Security. p.
  506–519. ASIA CCS '17, Association for Computing Machinery, New York, NY,
  USA (2017). \doi{10.1145/3052973.3053009},
  \url{https://doi.org/10.1145/3052973.3053009}

\bibitem{papernot2016distillation}
Papernot, N., McDaniel, P., Wu, X., Jha, S., Swami, A.: Distillation as a
  defense to adversarial perturbations against deep neural networks. In: 2016
  IEEE Symposium on Security and Privacy (SP). pp. 582--597. IEEE (2016)

\bibitem{DBLP:journals/corr/PapernotMG16}
Papernot, N., McDaniel, P.D., Goodfellow, I.J.: Transferability in machine
  learning: from phenomena to black-box attacks using adversarial samples. CoRR
   \textbf{abs/1605.07277} (2016), \url{http://arxiv.org/abs/1605.07277}

\bibitem{phan2021adversarial}
Phan, B., Mannan, F., Heide, F.: Adversarial imaging pipelines. In: Proceedings
  of the IEEE/CVF Conference on Computer Vision and Pattern Recognition. pp.
  16051--16061 (2021)

\bibitem{poursaeed2018generative}
Poursaeed, O., Katsman, I., Gao, B., Belongie, S.: Generative adversarial
  perturbations. In: Proceedings of the IEEE Conference on Computer Vision and
  Pattern Recognition. pp. 4422--4431 (2018)

\bibitem{prakash2018deflecting}
Prakash, A., Moran, N., Garber, S., DiLillo, A., Storer, J.: Deflecting
  adversarial attacks with pixel deflection. In: 2018 IEEE/CVF Conference on
  Computer Vision and Pattern Recognition (CVPR). IEEE (2018)

\bibitem{rauber2018foolbox}
Rauber, J., Brendel, W., Bethge, M.: Foolbox: A python toolbox to benchmark the
  robustness of machine learning models (2018)

\bibitem{ren2016faster}
Ren, S., He, K., Girshick, R., Sun, J.: Faster r-cnn: Towards real-time object
  detection with region proposal networks (2016)

\bibitem{samangouei2018defense}
Samangouei, P., Kabkab, M., Chellappa, R.: Defense-gan: Protecting classifiers
  against adversarial attacks using generative models. ICLR  (2018)

\bibitem{10.1109/TIP.2018.2872858}
Schwartz, E., Giryes, R., Bronstein, A.M.: Deepisp: Toward learning an
  end-to-end image processing pipeline  \textbf{28}(2),  912–923 (Feb 2019).
  \doi{10.1109/TIP.2018.2872858},
  \url{https://doi.org/10.1109/TIP.2018.2872858}

\bibitem{sen2020empir}
Sen, S., Ravindran, B., Raghunathan, A.: Empir: Ensembles of mixed precision
  deep networks for increased robustness against adversarial attacks. ICLR
  (2020)

\bibitem{shafahi2019adversarial}
Shafahi, A., Najibi, M., Ghiasi, A., Xu, Z., Dickerson, J., Studer, C., Davis,
  L.S., Taylor, G., Goldstein, T.: Adversarial training for free! In:
  Proceedings of the 33rd International Conference on Neural Information
  Processing Systems. pp. 3358--3369 (2019)

\bibitem{Shi2019CurlsW}
Shi, Y., Wang, S., Han, Y.: Curls \& whey: Boosting black-box adversarial
  attacks. 2019 IEEE/CVF Conference on Computer Vision and Pattern Recognition
  (CVPR) pp. 6512--6520 (2019)

\bibitem{stutz2020confidence}
Stutz, D., Hein, M., Schiele, B.: Confidence-calibrated adversarial training:
  Generalizing to unseen attacks. In: International Conference on Machine
  Learning. pp. 9155--9166. PMLR (2020)

\bibitem{szegedy2013intriguing}
Szegedy, C., Zaremba, W., Sutskever, I., Bruna, J., Erhan, D., Goodfellow, I.,
  Fergus, R.: Intriguing properties of neural networks. arXiv preprint
  arXiv:1312.6199  (2013)

\bibitem{szegedy2014intriguing}
Szegedy, C., Zaremba, W., Sutskever, I., Bruna, J., Erhan, D., Goodfellow, I.,
  Fergus, R.: Intriguing properties of neural networks (2014)

\bibitem{Tseng2021DeepCompoundOptics}
Tseng, E., Mosleh, A., Mannan, F., St-Arnaud, K., Sharma, A., Peng, Y., Braun,
  A., Nowrouzezahrai, D., Lalonde, J.F., Heide, F.: Differentiable compound
  optics and processing pipeline optimization for end-to-end camera design. ACM
  Transactions on Graphics (TOG)  \textbf{40}(4) (2021)

\bibitem{10.1145/3306346.3322996}
Tseng, E., Yu, F., Yang, Y., Mannan, F., Arnaud, K.S., Nowrouzezahrai, D.,
  Lalonde, J.F., Heide, F.: Hyperparameter optimization in black-box image
  processing using differentiable proxies  \textbf{38}(4) (Jul 2019).
  \doi{10.1145/3306346.3322996}, \url{https://doi.org/10.1145/3306346.3322996}

\bibitem{tseng2019hyperparameter}
Tseng, E., Yu, F., Yang, Y., Mannan, F., Arnaud, K.S., Nowrouzezahrai, D.,
  Lalonde, J.F., Heide, F.: Hyperparameter optimization in black-box image
  processing using differentiable proxies. ACM Trans. Graph.  \textbf{38}(4),
  27--1 (2019)

\bibitem{tsipras2019robustness}
Tsipras, D., Santurkar, S., Engstrom, L., Turner, A., Madry, A.: Robustness may
  be at odds with accuracy. In: International Conference on Learning
  Representations. No.~2019 (2019)

\bibitem{tu2019autozoom}
Tu, C.C., Ting, P., Chen, P.Y., Liu, S., Zhang, H., Yi, J., Hsieh, C.J., Cheng,
  S.M.: Autozoom: Autoencoder-based zeroth order optimization method for
  attacking black-box neural networks. In: Proceedings of the AAAI Conference
  on Artificial Intelligence. vol.~33, pp. 742--749 (2019)

\bibitem{wang2019bilateral}
Wang, J., Zhang, H.: Bilateral adversarial training: Towards fast training of
  more robust models against adversarial attacks. In: Proceedings of the
  IEEE/CVF International Conference on Computer Vision. pp. 6629--6638 (2019)

\bibitem{wang_simoncelli_bovik}
Wang, Z., Simoncelli, E., Bovik, A.: Multiscale structural similarity for image
  quality assessment. The Thrity-Seventh Asilomar Conference on Signals,
  Systems and amp; Computers, 2003 . \doi{10.1109/acssc.2003.1292216}

\bibitem{wong2020fast}
Wong, E., Rice, L., Kolter, J.Z.: Fast is better than free: Revisiting
  adversarial training. ICLR  (2020)

\bibitem{wu2020adversarial}
Wu, Y.H., Yuan, C.H., Wu, S.H.: Adversarial robustness via runtime masking and
  cleansing. In: International Conference on Machine Learning. pp.
  10399--10409. PMLR (2020)

\bibitem{xie2017mitigating}
Xie, C., Wang, J., Zhang, Z., Ren, Z., Yuille, A.: Mitigating adversarial
  effects through randomization. arXiv preprint arXiv:1711.01991  (2017)

\bibitem{xie2017adversarial}
Xie, C., Wang, J., Zhang, Z., Zhou, Y., Xie, L., Yuille, A.: Adversarial
  examples for semantic segmentation and object detection (2017)

\bibitem{xie2019feature}
Xie, C., Wu, Y., Maaten, L.v.d., Yuille, A.L., He, K.: Feature denoising for
  improving adversarial robustness. In: Proceedings of the IEEE Conference on
  Computer Vision and Pattern Recognition. pp. 501--509 (2019)

\bibitem{8953691}
Xu, X., Ma, Y., Sun, W.: Towards real scene super-resolution with raw images.
  In: 2019 IEEE/CVF Conference on Computer Vision and Pattern Recognition
  (CVPR). pp. 1723--1731 (2019). \doi{10.1109/CVPR.2019.00182}

\bibitem{xu2021exploiting}
Xu, X., Ma, Y., Sun, W., Yang, M.H.: Exploiting raw images for real-scene
  super-resolution. arXiv preprint arXiv:2102.01579  (2021)

\bibitem{yin2019gat}
Yin, X., Kolouri, S., Rohde, G.K.: Gat: Generative adversarial training for
  adversarial example detection and robust classification. In: International
  Conference on Learning Representations (2019)

\bibitem{Yu2021ReconfigISPRC}
Yu, K., Li, Z., Peng, Y., Loy, C.C., Gu, J.: Reconfigisp: Reconfigurable camera
  image processing pipeline. ArXiv  \textbf{abs/2109.04760} (2021)

\bibitem{zhang2016colorful}
Zhang, R., Isola, P., Efros, A.A.: Colorful image colorization. In: European
  conference on computer vision. pp. 649--666. Springer (2016)

\bibitem{zhang2019zoom}
Zhang, X., Chen, Q., Ng, R., Koltun, V.: Zoom to learn, learn to zoom. In:
  Proceedings of the IEEE/CVF Conference on Computer Vision and Pattern
  Recognition. pp. 3762--3770 (2019)

\bibitem{zheng2020efficient}
Zheng, H., Zhang, Z., Gu, J., Lee, H., Prakash, A.: Efficient adversarial
  training with transferable adversarial examples. In: Proceedings of the
  IEEE/CVF Conference on Computer Vision and Pattern Recognition. pp.
  1181--1190 (2020)

\end{thebibliography}
\end{document}